\documentclass[10pt]{article}
\usepackage[a4paper, total={6in, 8in}]{geometry}
\usepackage{graphicx}
\usepackage{subfig}
\usepackage{lmodern}
\usepackage[latin9]{inputenc}

\usepackage{chngcntr}

\usepackage{float}
\usepackage{multirow}
\usepackage{algorithm}
\usepackage{algorithmic}
\usepackage{amsmath}
\usepackage{xcolor}
\usepackage{rotating}
\usepackage{enumitem}
\usepackage{longtable}
\usepackage{chngcntr}
\usepackage{lscape}
\usepackage{todonotes}   
\usetikzlibrary{arrows}
\usetikzlibrary{shapes}
\usetikzlibrary{positioning}
\usepackage{authblk}
\usepackage{lineno}
\newcommand{\mymk}[1]{%
  \tikz[baseline=(char.base)]\node[anchor=south west, draw,rectangle, rounded corners, inner sep=1pt, minimum size=3mm,
    text height=1.2mm](char){\ensuremath{#1}} ;}

\begin{document}


\title{Scalarizing Functions in Bayesian Multiobjective Optimization}
\author{Tinkle Chugh}
\affil{Department of Computer Science, University of Exeter, UK}
\date{} 
\maketitle


\begin{abstract}
Scalarizing functions have been widely used to convert a multiobjective optimization problem into a single objective optimization problem. However, their use in solving (computationally) expensive multi- and many-objective optimization problems in Bayesian multiobjective optimization is scarce. Scalarizing functions can play a crucial role on the quality and number of evaluations required when doing the optimization. In this article, we study and review 15 different scalarizing functions in the framework of Bayesian multiobjective optimization and build Gaussian process models (as surrogates, metamodels or emulators) on them. We use expected improvement as infill criterion (or acquisition function) to update the models. In particular, we compare different scalarizing functions and analyze their performance on several benchmark problems with different number of objectives to be optimized. The review and experiments on different functions provide useful insights when using and selecting a scalarizing function when using a Bayesian multiobjective optimization method. 

\textbf{Keywords:} metamodelling, machine learning, multiple criteria decision making, Pareto optimality, computational cost, Bayesian optimization
\end{abstract}
\section{INTRODUCTION}
Bayesian multiobjective optimization methods have been widely used for solving (computationally) expensive multiobjective optimization problems (MOPs) and in the last few years, many articles have been published. A survey of many Bayesian and non-Bayesian methods with their advantages and limitations is provided in \cite{Chugh2018,Jin2005,Jin2011,Richard2016, Deb2018,Horn2015}. 



\begin{figure*}
\centering
\includegraphics[scale=0.5]{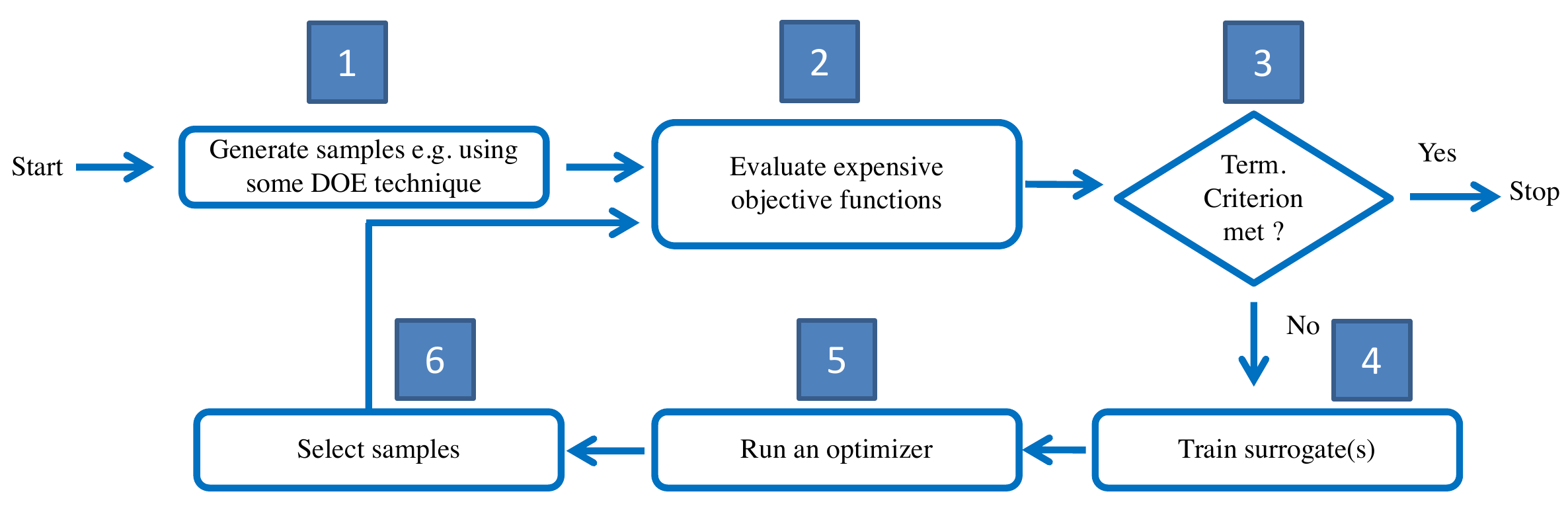}
\caption{\label{fig:surrogate} A generic framework for a surrogate-assisted optimization method}
\end{figure*}

A generic framework of a Bayesian optimization method is shown in Figure \ref{fig:surrogate}, where the steps are mentioned inside squares. 
In the first step, a set of samples is generated e.g.\ using a design of experiment technique (DOE) like \ Latin hypercube sampling \cite{Mckay2000}. These samples are then evaluated with expensive objective functions in the second step. A termination criterion (typically maximum number of expensive function evaluations) is checked in the third step. If the termination criterion is met, nondominated solutions from all evaluated solutions (i.e.\ from expensive evaluations) are used as the final solutions. Otherwise, surrogate models are built using evaluated solutions in the fourth step. An optimizer e.g.\ an evolutionary algorithm is then used with the models in the fifth step to find promising samples by using an appropriate infilling criterion (or updating criterion or acquisition function). In the sixth step, a fixed number of samples generated with the optimizer is selected which are then evaluated with expensive objective functions.


All six steps mentioned above are important in the performance of a Bayesian optimization method. In this article, our focus is on the fourth step i.e.\ building or training surrogates. Once, the samples are evaluated with expensive objective functions in the second step, there can be different ways to build surrogates to be used in a Bayesian optimization method. The most common way is to build surrogate for each objective function \cite{Chugh2016a,Ponweiser2008a,Zhang2010,Lim2010}. For instance, in \cite{Chugh2016a, Zhang2010, Ponweiser2008a}, Gaussian process models are used for different objective functions. Another way is to build a surrogate for a scalarizing function after converting multiobjective optimization problem into a single objective optimization problem \cite{Knowles2006,Alma2017}, which is also the focus of this article. There are other approaches for building surrogates like classification of solutions into different ranks or classes \cite{CSEA2018,Loshchilov2009,Loshchilov2010,Pilat2011b,Bandaru2014}. For details on other ways of building surrogates, see \cite{Chugh2018,Richard2016}. Note that scalarizing function can also be used after building models on each objective function e.g.\ as in \cite{Zhang2010, Yew_Soon2012} to do a local search. In this article, we focus on building surrogate after converting a multiobjective optimization problem into a single objective by using a scalarizing function. 

There are two main advantages to make a single surrogate in solving an expensive MOP. The first one is that only one surrogate is used in the solution process instead of multiple surrogates, which reduces the computational burden e.g.\ the training time especially in many-objective (usually more than three objectives) optimization problems. The second advantage is that one can use an infill criterion proposed for single-objective optimization problems which also reduces the computational complexity. In the literature, a little attention has been paid in using scalarizing functions for building surrogates on them and only a few studies exist in the literature. For instance, in \cite{Knowles2006}, an augmented achievement scalarizing function (AASF) was used and in \cite{Alma2017}, hypervolume improvement, dominance rank and minimum signed distance were used (more details are provided in the Section \ref{sec:scf_functions}). 


It is worth important to be pointed out that several studies exist \cite{Coello2017,Ishibuchi2016,Coello2017b,Miettinen2002,Miettinen1999} on using scalarizing functions without using them in a Bayesian optimization method. In this article, we study 15 different scalarizing functions to build surrogates on them in solving (computationally) expensive MOPs and particularly, focus on answering the following research questions:
\begin{enumerate}
\item are different scalarizing functions perform different to each other under the same framework?
\item are surrogate models built on a given scalarizing function sensitive to the number of objectives ?

\end{enumerate} 

To answer the questions above, we embed different scalarizing functions into the framework of efficient global optimization (EGO) \cite{Jones1998}. The algorithm uses the expected improvement criterion for updating the surrogates build on the scalarizing function. We then test the method with different scalarizing functions on several benchmark problems with different numbers of objectives. We analyze the accuracy and uncertainty provided by Gaussian process models on different scalarizing functions with the different number of objectives.

The rest of the article is organized as follows. In the next section, we provide a literature survey on using scalarizing functions with and without building surrogates on them. In Section 3, we give a brief introduction to different functions with their mathematical formulations and merits and demerits. In Section 4, numerical experiments are conducted to answer the research questions mentioned above. We finally conclude and mention the future research directions in Section 5. 

\section{Related work}
Use of scalarizing functions has been a topic of interest in different communities. For instance, in Multiple Criteria Decision Making (MCDM) problems, several scalarizing functions have been used to incorporate a decision maker's preferences \cite{Miettinen1999,Hwang1979,Steuer1986, Miettinen2002}. In evolutionary multiobjective optimization (EMO) algorithms especially decomposition based, utilization of scalarizing functions became popular in the last few years. Many studies and algorithms exist in the literature utilizing different functions for solving multi- and many-objectives optimization problems. Some of the well-known decomposition based EMO algorithms utilizing different scalarizing functions are nondominated-sorting genetic algorithm III (NSGA-III) \cite{Deb2014,Deb2014b}, multiobjective optimization based on decomposition (MOEA/D) \cite{Zhang2007} and its numerous versions \cite{Survey_MOEA/D} and reference vector guided evolutionary algorithm (RVEA) \cite{Ran2016}. For more details on the working principle of decomposition based algorithms, see surveys on many-objective optimization evolutionary algorithms \cite{Ishibuchi2008b,Li2015}.



In \cite{Hisao2010}, two different scalarizing functions weighted sum and augmented Chebyshev was used adaptively in the solution process in the framework of MOEA/D by using a multi-grid scheme. The proposed idea was tested on a knapsack problem with four and six objectives and performed better than the original version of MOEA/D. The authors extended their work in \cite{Hisao2016}, and modified the penalty boundary intersection (PBI) function in MOEA/D to handle different kinds of Pareto front. Two modified PBI functions called two-level PBI and quadratic PBI were tested in the MOEA/D framework and the algorithm with two new scalarizing functions performed better than the original version.

A detailed study on different scalarizing functions and their corresponding parameters was conducted in \cite{Derbel2014}. Instead of proposing a new algorithm, the authors showed the performance of the different scalarizing function in a simple (1+$\lambda$)-evolutionary algorithm \cite{Derbel2014} on bi-objective optimization problems. In \cite{Coello2017}, the performance of 15 different scalarizing functions was tested in MOEA/D \cite{Zhang2007} and MOMBI-II \cite{Coello2015} algorithms. The authors used a tool called EVOCA \cite{Riff2013} to tune the parameter values in different functions. Two algorithms MOEA/D and MOMBI-II \cite{Coello2015} with different scalarizing functions were tested on Lame Supersphere test problems \cite{Emmerich2007}. The authors found out that the performance of two algorithms depends on the choice of scalarizing functions. 


In \cite{Coello2017b}, a hyper-heuristic was used to rank different scalarizing functions with a measure called s-energy \cite{Hardin2004}. The proposed algorithm was tested on ZDT, DTLZ and WFG benchmark problems with 2-10 objectives and compared with MOMBI-II, MOEA/D and NSGA-III and found better results in most of the instances.  In \cite{Jiang2018}, two new scalarizing functions called the multiplicative scalarizing function (MSF) and penalty-based scalarizing function (PSF) was used in MOEA/D-DE \cite{Zhang2017}. The proposed scalarizing functions performed better than penalty boundary intersection, Chebyshev and weighted sum scalarizing functions. 

The first algorithm in Bayesian multiobjective optimization method using a scalarizing function and building a surrogate on it was proposed in \cite{Knowles2006} and known as ParEGO (for Pareto based efficient global optimization). A Gaussian process model \cite{Rasmussen2006} was used as a surrogate of the scalarizing function. In the algorithm, a set of reference vectors was uniformly generated in the objective space using simplex lattice-design method \cite{Cornell2011}. In each iteration, the algorithm randomly selected a reference vector among a set of vectors, which was then used in the Chebyshev function to build the surrogate. The algorithm was compared with NSGA-II \cite{Deb2002} and performed significantly better. 

In a recent study in \cite{Alma2017}, three scalarizing functions, hypervolume improvement, dominance ranking, and minimum signed distance were proposed in solving expensive MOP. These scalarizing functions do not use reference vectors and preserve the dominance relationship. The authors also used the framework of EGO and compared with SMS-EGO \cite{Ponweiser2008a} and ParEGO (i.e.\ with Chebyshev scalarizing function). The results outperformed the ParEGO algorithm and performed similarly to SMS-EGO in some cases. However, the reason for the better performance of their approach comparison to ParEGO was not mentioned. 







\section{Scalarizing functions}
\label{sec:scf_functions}
In this section, we summarize 15 different scalarizing functions studies in this work with their merits and demerits. All these functions have already been explained in details in the literature \cite{Coello2007, Miettinen1999}. Therefore, we provide a brief summary of these functions.
Before going into description of scalarizing function, we define a multiobjective optimization problem (MOP) as:

\begin{equation}
\begin{gathered}
\mbox{minimize }  {\left\{f_1(x),\ldots,f_k(x)\right\}} \\
\mbox{subject to} \  x \in S 
\end{gathered}
\label{MOOP}
\end{equation}
with $k (\geq2)$ objective functions $f_i(x)$: \textit{S}$\rightarrow\Re^n$. The vector of objective function values is denoted by $f(x)=(f_1(x),\ldots,f_k(x))^T$. The (nonempty) feasible space $S$ is a subset of the decision space $\Re^n$ and consists of decision vectors $x=(x_1,\ldots,x_n)^T$ that satisfy all the constraints. Different scalarizing functions are defined as follows:

\begin{enumerate}[leftmargin=*]
\item \textbf{Weighted sum (WS)}: The weighted sum combines different objectives linearly and has been widely used \cite{Miettinen1999}. It converts the a MOP into a single-objective optimization problems as:
\begin{equation}
  g = \sum_{i=1}^{k}w_if_i.
\end{equation}
One major limitation in using WS is that it cannot find solutions in non-convex parts of the Pareto front.

\item \textbf{Exponential weighted criterion (EWC)}: EWC was first used in \cite{Athan1996} to overcome the limitations of WS:
\begin{equation}
g = \sum_{i=1}^{k}\exp(pw_i-1)\exp(pf_i).
\end{equation}
As mentioned in \cite{Coello2007}, the performance of the function depends on the value of $p$ and usually a large value of $p$ is needed. 

\item \textbf{Weighted power (WPO)}: The weighted power defined below can find solutions when the Pareto front is not convex. However, it also depends on the parameter $p$ as in EWC. 
\begin{equation}
 g = \sum_{i=1}^{k}w_i(f_i)^p.
\end{equation}

\item \textbf{Weighted norm (WN)}: The weighted norm (or $L_p$ metric) is a generalized form of the weighted sum:
\begin{equation}
 g = (\sum_{i=1}^{k}w_i|f_i|^p)^{1/p}.
\end{equation}
The weighted norm has been used in MOEA/D \cite{Zhang2007}. Likewise EWC and WPO, its performance also depends on the choice of the $p$ parameter. 

\item \textbf{Weighted product (WPR)}: The weighted power is defined as:
\begin{equation}
 g = \prod_{i=1}^{k}(f_i)^{w_i}.
\end{equation}
It is also called as product of powers \cite{Coello2007} and can find solutions in non-convex parts of the Pareto front. However, like EWC, WPO and WN, its performance depends on the value of $p$.


\item \textbf{Chebyshev function (TCH)}:  The Chebyshev function can be derived from the WN function with $p=\infty$. This function has been used in many EMO algorithms such as MOEA/D \cite{Zhang2007} and its versions \cite{Survey_MOEA/D}:
\begin{equation}
 g = \max_i[w_i|f_i - z_i^*|],
\end{equation}
where $z_i^*$ is the ideal or utopian objective vector. 
In this work, we normalize the expensive objective function values in the range [0,1] before building the model. Therefore, $z_i^*$ is a vector of zeros. 


\item \textbf{Augmented Chebyshev (ATCH)}: In \cite{Steuer1986}, it was suggested that weakly Pareto optimal solutions can be avoided by adding an augmented term to TCH:
\begin{equation}
 g = \max_i[w_i|f_i - z_i^*|] + \alpha\sum_{i=1}^{k}|f_i - z_i^*|
\end{equation}
Moreover, this function was the first to be used as a surrogate in \cite{Knowles2006}.

\item \textbf{Modified Chebyshev (MTCH)}: A slightly modified form of MTCH was used in \cite{Kaliszewski1987}:
\begin{equation}
 g = \max_i\big[ w_i(|f_i - z_i^*| + \alpha\sum_{i=1}^{k}|f_i - z_i^*|)\big]
\end{equation}
As mentioned in \cite{Miettinen1999} that main different between ATCH and MTCH is the slope to avoid weakly Pareto optimal solutions.






\item \textbf{Penalty boundary intersection (PBI)}: The PBI function was first used in MOEA/D and used as the selection criterion for balancing convergence and diversity:
\begin{equation}
 g = d_1 + \theta d_2,
\end{equation}where $d_1 = |\mathbf{f} \cdot \frac{\mathbf{w}}{\| \mathbf{w} \|}|$ and $d_2 = \|\mathbf{f} - d_1 \frac{\mathbf{w}}{\| \mathbf{w} \|}\|$. The PBI function has been widely used in EMO algorithms \cite{Survey_MOEA/D}. However, as shown in  \cite{Ishibuchi2010} its performance is effected by the $\theta$ parameter.

\item \textbf{Inverted penalty boundary intersection (IPBI)}: To enhance the diversity of solutions, IPBI was proposed in \cite{Sato2014}:
\begin{equation}
 g = \theta d_2 - d_1,
\end{equation}
where $d_1 = |\mathbf{f}^{'} \cdot \frac{\mathbf{w}}{\| \mathbf{w} \|}|$ and $d_2 = \|\mathbf{f}^{'} - d_1 \frac{\mathbf{w}}{\| \mathbf{w} \|}\|$ and $\mathbf{f}^{'} = \mathbf{z}^{nadir} - \mathbf{f}$. In this work, we considered the vector of worst expensive objective function values as $\mathbf{z^nadir}$. 

\item \textbf{Quadratic PBI  (QPBI)}: Recently, an enhanced version of PBI function was proposed in \cite{Hisao2016}:
\begin{equation}
g = d_1 + \theta d_2 \frac{d_2}{d^*},
\end{equation}
where $d_1$ and $d_2$ are same as in PBI and $d^*$ is an adaptive parameter and defined as: 
\begin{equation}
d^* = \alpha \frac{1}{H} \frac{1}{k} \sum (z^{nadir} - z^{ideal})
\end{equation}
where $\alpha$ is a pre-defined parameter and $H$ is a parameter used in generating the reference (or weight vectors) in decomposition based EMO algorithms. For more details about these parameters, see \cite{Zhang2007,Ran2016}. 



\item \textbf{Angle penalized distance (APD)}: The APD function is a recently proposed scalarizing function and used as the selection criterion in reference vector guided many-objective evolutionary algorithm (RVEA) \cite{Ran2016}:
\begin{equation}
g = 1 + P(\theta) \cdot \|\mathbf{f}\|,
\end{equation}
where $P(\theta) = k (\frac{FE}{FE^{max}})^{\alpha} \frac{\theta}{\gamma}$, $FE$ is the number of expensive function evaluations at the current iteration and $FE^{max}$ is the maximum number of expensive function evaluations. The angle between an objective vector $\mathbf{f}$ and the reference vector to which it is assigned is represented by $\theta$ and the minimum of all angles between a reference vector selected and other reference vectors is represented by $\gamma$. This function adaptively balances the convergence and diversity based on the maximum number of function evaluations. The performance of the function depends on the value of $\alpha$ and $FE^{max}$.




\item \textbf{Hypervolume improvement (HypI)}: The HypI is a recently proposed scalarizing function in a surrogate-based algorithm \cite{Alma2017}. In contrast to its name, this scalarizing function is designed in such a way that it uses the Pareto dominance. Given a set of solutions $X$, a nondominated sorting is performed to find fronts of different ranks as in \cite{Deb2002}. Let the different fronts of ranks $1,2, \ldots$ are denoted by $P_1, P_2, \ldots$ and the hypervolume of a front $P_k$ given a reference point $r$ is denoted by $H(P_k)$. Then hypervolume improvement a solution $x \in X$ belongs to the front $P_k$ is then given by:
\begin{equation}
g = H(x \cup P_{k+1}).
\end{equation}
In this way, the Pareto dominance is preserved when calculating the hypervolume contributions of solutions. For problems with many-objectives, where all solutions are non-dominated i.e.\ belong to only one front, this scalarizing function might not be suitable. In other words, if all solutions are nondominated, their HypI values will be similar and the algorithm embedding the model will not be able to solve problems with many-objectives. Moreover, the performance of the function can be sensitive to the reference point in calculating the hypervolume.

\item \textbf{Dominance ranking (DomRank)}: 
This function is also recently proposed in \cite{Alma2017} and assigns fitness values based on the ranks of different solutions as done in the MOGA algorithm \cite{Fonseca1993}. In \cite{Fonseca1993}, a solution is assigned a rank as: rank(x)= 1+$p$, where $p$ is number of solutions dominating x. Similarly, in \cite{Alma2017}, given a set of solutions (with expensive evaluations) $X$ the fitness of a solution $x$ is:
 \begin{equation}
g = 1 - \frac{rank(x)-1}{|X|-1}.
\end{equation}
For instance, the rank of a solution $x'$ dominated by all other solutions would be $rank(x') = 1 + |X| -1$, and therefore, the fitness of solution would be $g(x') = 1 - \frac{1 + |X| - 1 -1}{|X|-1} = 0$. Similarly, the rank of solutions belong the first front will be 1. This function is maximized to find samples for training the surrogates. 
Similar to HypI, this function also uses Pareto dominance and might not be suitable for problems with many-objectives. 

\item \textbf{Minimum Signed Distance (MSD) \cite{Alma2017}}: The MSD function is proposed in \cite{Alma2017} and defined as:
\begin{equation}
g = \min d(x', x)
\end{equation}
where $x'$ are the solutions belong the first front (or rank one solutions) and $d(x',x) = \sum_{i=1}^k f_i(x') - f_i(x)$. For instance, if a solution a dominates another solution b, then $g(a,x') > g(b,x')$. 

\end{enumerate}



\section{Numerical experiments}
This section provides a comparison of different functions and analysis when building surrogates on them. As mentioned we used the framework of EGO \cite{Jones1998} and built Gaussian process as surrogates on the scalarizing functions. The expected improvement criterion in EGO is maximized with a genetic algorithm for selecting samples when updating the surrogates. 
\subsection{Performance of different scalarizing functions}
To compare the performance of different scalarizing functions, we used DTLZ \cite{Deb2005} and WFG \cite{Huband2005} problems with 2, 3, 5 and 10 number of objectives. In DTLZ suite, the number of variables is kept to $k$+5-1, where $k$ is the number of objectives. For the WFG suite, the number of variables is defined by position ($d$) and distance ($l$) parameters. We used the number of parameters as suggested in \cite{Huband2005}. For two objectives, $d$ is set to four and $2\times(k-1)$ for the rest of the objectives. The distance parameter is set to four for all objectives. To be summarized, the number of variables ($n$) for DTLZ and WFG suites is given in Table \ref{tab:no_variables_WFG}:
\begin{table}
\centering
\caption{\label{tab:no_variables_WFG}Number of variables ($n$) for WFG and DTLZ suites}
\begin{tabular}{|c|c|c|c|c|}
\hline
   & \multicolumn{3}{c|}{WFG} & DTLZ \\ \hline
$k$  & $d$       & $l$     & $n$      & $n$    \\ \hline
2  & 4       & 4     & 8      & 6    \\ \hline
3  & 4       & 4     & 8      & 7    \\ \hline
5  & 8       & 4     & 12     & 9    \\ \hline
10 & 18      & 4     & 22     & 14   \\ \hline
\end{tabular}
\end{table}

There are several parameters in different scalarizing functions and we used the recommended values from the respective articles. Different parameter values used are provided in Table \ref{tab:parameters}:
\begin{table}
    \centering
    \caption{\label{tab:parameters} Parameters values used in different scalarizing functions}
    \begin{tabular}{|c|c|} \hline
    Scalarizing function     & Parameter value \\ \hline
       EWC  & $p$ = 100\\ 
       WPO & $p$ = 3\\
       WN &  $p$ = 0.5 \\
       ATCH & $\alpha$ = 0.0001\\
       MTCH & $\alpha$ = 0.0001 \\
       PBI & $\theta$ = 5\\
       IPBI & $\theta$ = 5\\
       QPBI & $\theta$ = 1\\
       APD & $\alpha$ = 2 \\
       HypI & $r$ (In table \ref{tab:ref_points} for DTLZ problems)\\ \hline
    \end{tabular}
\end{table}
\begin{table}
    \centering
     \caption{\label{tab:ref_points}Reference point in calculating hypevolumes in HypI and as performance measure: }
    \begin{tabular}{|c|c|} 
    \hline
    Problem & reference point \\ \hline
       DTLZ1  & 400 $\times$ (1,k) \\
DTLZ2         & 1.5 $\times$ (1,k) \\
DTLZ3 & 900 $\times$ (1,k) \\
DTLZ4 & 2 $\times$(1,k) \\
DTLZ5 & 1.5 $\times$ (1,k)  \\
DTLZ6 & 6 $\times$ (1,k) \\
DTLZ7 & 5 $\times$ (1,k)  \\
\hline
\end{tabular}
\end{table}
For WFG problems, the reference point $r$ in HyPI is used as $2 \times z^{nadir}$. The vector $z^{nadir}$ contains the maximum objective function values in the Pareto front of the given problem. We ran 21 independent runs for each scalarizing function with 300 maximum number of expensive function evaluations. We show the performance of different functions with inverted generational distance (IGD) and hypervolume. The mean IGD values with standard deviation (in parentheses) for DTLZ problems are provided in Table \ref{tab:IGD_DTLZ} and other results including IGD and hypervolume on WFG are in the supplementary material\footnote{available from https://github.com/tichugh/SCF}. To compare the results of different functions, we used the Wilcoxon rank sum test and used the Gaussian (or RBF) kernel when using the Gaussian process model. 



\begin{figure}
    \centering
    \includegraphics[scale=0.4]{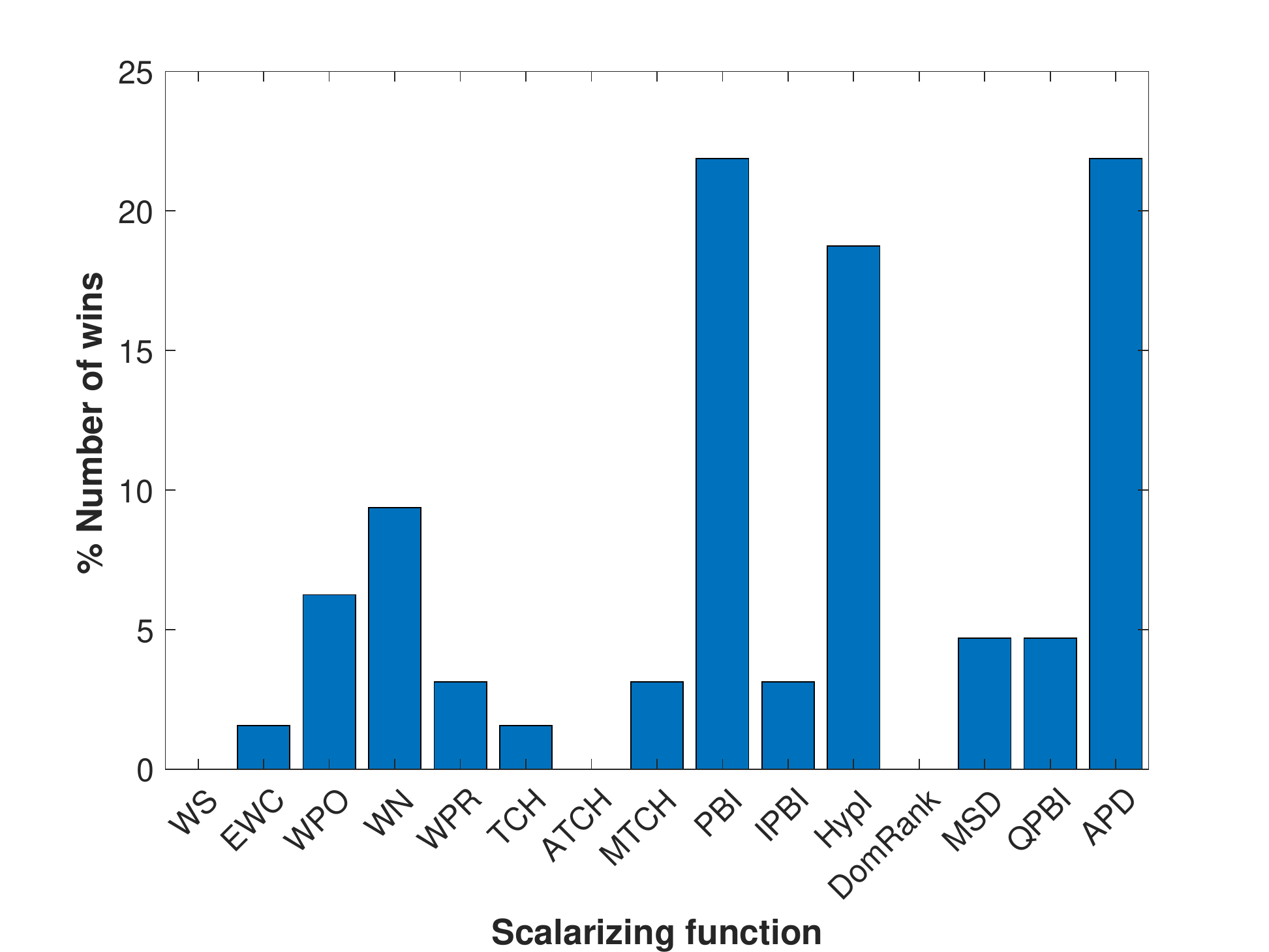}
    \caption{\label{fig:bar_graph}Number of wins of different scalarizing functions}
\end{figure}

In tables, the values statistically similar to the best value are in bold, the best value is in the circle and the worst value is underlined. As can be seen, the performance of different functions varies with different problems. Moreover, we provide in how many instances, a particular scalarizing function has the best IGD value in Figure \ref{fig:bar_graph}. The total number of instances tested were (7 + 9) $\times$ 4 = 64 i.e.\ (number of DTLZ problems + number of WFG problems) $\times$ number of cases with respect to objectives. 

As can be seen from tables and the figure, three functions PBI, HypI, and APD outperformed other functions. Moreover, WS, ATCH, and DomRank never had the best IGD value in any of 64 instances. Note that, all these three functions have statistically similar values to the best IGD value in many cases as can be seen from tables. Another interesting observation from the results is that three different scalarizing functions ATCH, DomRank and MSD used in the literature for building surrogates on them were not in the top list. On the other hand, scalarizing functions used as the selection criterion in EMO algorithms e.g.\ APD and PBI performed significantly better. These results indicate that it is not straightforward to select a particular scalarizing function. However, one can analyze the fitness landscape of different functions after building surrogates. Moreover, this landscape can be sensitive to the number of objectives and the details are provided in the next subsection. 


\begin{table*} 
\scalebox{0.6}{
\begin{tabular}{|l|l|c|c|c|c|c|c|c|c|} 
\hline 
Problem & k& WS& EWC& WPO& WN& WPR& TCH& ATCH& MTCH \\ 
 \hline
 & 2& {\color{black}{\textbf{42.26}}} (8.13)& {\color{black}{\textbf{46.28}}} (16.52)& {\color{black}{\textbf{37.95}}} (20.37)& {\color{black}{\textbf{41.54}}} (7.98)& {\color{black}{\textbf{38.38}}} (9.24)& {\color{black}{\textbf{41.15}}} (7.59)& {\color{black}{\textbf{38.86}}} (5.55)& {\color{black}{\textbf{38.98}}} (7.00)\\
 DTLZ1& 3& {\color{black}{\textbf{41.55}}} (5.51)& {\color{black}{\textbf{41.03}}} (14.44)& {\color{black}{\textbf{33.07}}} (7.59)& {\color{black}{\textbf{42.60}}} (5.81)& {\color{black}{\textbf{34.86}}} (5.83)& {\color{black}{\textbf{39.33}}} (3.66)& {\color{black}{\textbf{39.81}}} (2.91)& {\color{black}{\textbf{40.32}}} (2.86)\\
 & 5& 38.40 (5.30)& 34.56 (11.79)& 29.82 (8.37)& 38.85 (5.24)& 37.24 (10.61)& 36.92 (5.07)& 38.88 (5.72)& 38.23 (3.84)\\
 & 10& {\color{black}{\textbf{38.65}}} (12.18)& {\color{black}{\textbf{39.07}}} (15.04)& {\color{black}{\textbf{34.51}}} (9.55)& {\color{black}{\textbf{39.29}}} (13.78)& {\color{black}{\textbf{31.14}}} (11.38)& {\color{black}{\textbf{35.91}}} (12.20)& {\color{black}{\textbf{37.54}}} (10.52)& {\color{black}{\textbf{38.09}}} (9.53)\\\hline
 & 2& 0.19 (0.03)& 0.14 (0.03)& {\color{black}{\textbf{0.05}}} (0.01)& {\color{black}{\underline{0.19}}} (0.03)& 0.19 (0.03)& {\color{black}{\textbf{0.04}}} (0.00)& {\color{black}{\textbf{0.04}}} (0.00)& {\color{black}\mymk{\textbf{0.04}}} (0.00)\\
 DTLZ2& 3& 0.26 (0.02)& 0.21 (0.02)& {\color{black}{\textbf{0.14}}} (0.01)& {\color{black}{\underline{0.27}}} (0.02)& 0.25 (0.01)& 0.16 (0.01)& 0.16 (0.01)& 0.17 (0.02)\\
 & 5& 0.41 (0.02)& 0.39 (0.04)& 0.32 (0.01)& 0.40 (0.02)& 0.37 (0.02)& 0.40 (0.01)& 0.40 (0.02)& 0.40 (0.01)\\
 & 10& 0.81 (0.04)& 0.82 (0.03)& {\color{black}{\textbf{0.76}}} (0.02)& 0.80 (0.05)& 0.83 (0.04)& 0.80 (0.03)& 0.81 (0.04)& 0.81 (0.03)\\\hline
 & 2& {\color{black}{\textbf{93.74}}} (10.84)& {\color{black}{\textbf{123.41}}} (43.55)& {\color{black}{\textbf{94.69}}} (20.13)& {\color{black}{\textbf{88.45}}} (12.71)& {\color{black}\mymk{\textbf{87.19}}} (8.14)& {\color{black}{\textbf{88.54}}} (14.42)& {\color{black}{\textbf{89.96}}} (12.05)& {\color{black}{\textbf{92.26}}} (12.10)\\
 DTLZ3& 3& {\color{black}{\textbf{99.90}}} (8.77)& 115.48 (20.65)& {\color{black}{\textbf{90.15}}} (25.95)& {\color{black}{\textbf{90.38}}} (8.50)& {\color{black}{\textbf{89.41}}} (14.54)& {\color{black}{\textbf{99.88}}} (6.58)& {\color{black}{\textbf{97.63}}} (8.99)& {\color{black}{\textbf{93.69}}} (12.77)\\
 & 5& {\color{black}{\textbf{93.10}}} (13.14)& 137.71 (30.78)& {\color{black}{\textbf{105.19}}} (32.37)& {\color{black}{\textbf{100.31}}} (11.86)& {\color{black}{\textbf{93.92}}} (28.96)& {\color{black}{\textbf{96.68}}} (15.43)& {\color{black}{\textbf{95.11}}} (17.12)& {\color{black}{\textbf{95.96}}} (11.63)\\
 & 10& {\color{black}{\textbf{101.40}}} (15.91)& {\color{black}{\textbf{116.94}}} (30.53)& {\color{black}{\textbf{103.88}}} (22.65)& {\color{black}{\textbf{109.04}}} (25.48)& {\color{black}{\textbf{117.87}}} (32.94)& {\color{black}{\textbf{90.99}}} (21.19)& {\color{black}{\textbf{103.90}}} (14.91)& {\color{black}{\textbf{93.17}}} (11.40)\\\hline
 & 2& 0.34 (0.04)& 0.41 (0.16)& {\color{black}{\textbf{0.34}}} (0.03)& 0.38 (0.09)& {\color{black}{\underline{0.45}}} (0.09)& {\color{black}{\textbf{0.32}}} (0.05)& {\color{black}{\textbf{0.33}}} (0.03)& {\color{black}{\textbf{0.33}}} (0.03)\\
 DTLZ4& 3& {\color{black}{\textbf{0.50}}} (0.05)& 0.54 (0.08)& {\color{black}{\textbf{0.46}}} (0.03)& {\color{black}{\underline{0.56}}} (0.05)& 0.54 (0.08)& {\color{black}{\textbf{0.47}}} (0.03)& {\color{black}{\textbf{0.48}}} (0.04)& {\color{black}{\textbf{0.47}}} (0.04)\\
 & 5& 0.64 (0.03)& 0.67 (0.06)& {\color{black}{\textbf{0.57}}} (0.02)& 0.64 (0.03)& 0.69 (0.04)& 0.64 (0.03)& 0.65 (0.04)& 0.65 (0.04)\\
 & 10& {\color{black}{\textbf{0.76}}} (0.03)& 0.80 (0.03)& {\color{black}{\textbf{0.75}}} (0.03)& {\color{black}{\textbf{0.77}}} (0.02)& 0.82 (0.02)& {\color{black}{\textbf{0.75}}} (0.03)& {\color{black}{\textbf{0.76}}} (0.02)& {\color{black}{\textbf{0.76}}} (0.03)\\\hline
 & 2& 0.21 (0.03)& 0.14 (0.02)& {\color{black}{\textbf{0.05}}} (0.01)& {\color{black}{\underline{0.21}}} (0.03)& 0.20 (0.03)& {\color{black}{\textbf{0.04}}} (0.01)& {\color{black}{\textbf{0.04}}} (0.01)& {\color{black}{\textbf{0.05}}} (0.00)\\
 DTLZ5& 3& {\color{black}{\underline{0.18}}} (0.02)& 0.13 (0.02)& {\color{black}\mymk{\textbf{0.03}}} (0.01)& 0.18 (0.02)& 0.18 (0.02)& 0.06 (0.01)& 0.07 (0.01)& {\color{black}{\textbf{0.06}}} (0.01)\\
 & 5& 0.18 (0.03)& 0.12 (0.02)& {\color{black}{\textbf{0.04}}} (0.01)& {\color{black}{\underline{0.18}}} (0.03)& 0.17 (0.03)& 0.09 (0.02)& 0.08 (0.01)& 0.09 (0.02)\\
 & 10& 0.15 (0.02)& 0.11 (0.01)& {\color{black}{\textbf{0.06}}} (0.01)& 0.15 (0.02)& {\color{black}{\underline{0.15}}} (0.02)& 0.14 (0.02)& 0.14 (0.03)& 0.14 (0.03)\\\hline
 & 2& {\color{black}{\textbf{0.41}}} (0.03)& {\color{black}{\underline{3.67}}} (0.32)& {\color{black}\mymk{\textbf{0.19}}} (0.04)& {\color{black}{\textbf{0.40}}} (0.02)& 0.62 (0.18)& {\color{black}{\textbf{0.36}}} (0.08)& {\color{black}{\textbf{0.37}}} (0.09)& {\color{black}{\textbf{0.39}}} (0.09)\\
 DTLZ6& 3& {\color{black}{\textbf{0.46}}} (0.07)& {\color{black}{\underline{3.58}}} (0.24)& {\color{black}{\textbf{0.49}}} (0.14)& {\color{black}{\textbf{0.53}}} (0.11)& {\color{black}{\textbf{0.82}}} (0.19)& {\color{black}{\textbf{0.42}}} (0.10)& {\color{black}{\textbf{0.46}}} (0.07)& {\color{black}{\textbf{0.44}}} (0.06)\\
 & 5& {\color{black}{\textbf{0.74}}} (0.12)& {\color{black}{\underline{3.71}}} (0.26)& 1.27 (0.42)& {\color{black}{\textbf{0.75}}} (0.15)& 1.19 (0.32)& {\color{black}{\textbf{0.85}}} (0.13)& {\color{black}{\textbf{0.75}}} (0.18)& {\color{black}{\textbf{0.88}}} (0.18)\\
 & 10& 1.71 (0.50)& 3.51 (0.33)& 1.62 (0.46)& 2.26 (0.59)& 2.99 (0.50)& 1.57 (0.39)& 1.68 (0.29)& 1.48 (0.32)\\\hline
 & 2& {\color{black}{\textbf{0.22}}} (0.02)& {\color{black}{\underline{3.27}}} (0.89)& {\color{black}{\textbf{0.30}}} (0.05)& {\color{black}{\textbf{0.24}}} (0.08)& {\color{black}{\textbf{0.35}}} (0.03)& {\color{black}{\textbf{0.25}}} (0.02)& {\color{black}{\textbf{0.25}}} (0.03)& {\color{black}{\textbf{0.25}}} (0.01)\\
 DTLZ7& 3& {\color{black}{\textbf{0.47}}} (0.05)& {\color{black}{\underline{5.10}}} (1.16)& {\color{black}{\textbf{0.69}}} (0.13)& {\color{black}{\textbf{0.56}}} (0.04)& {\color{black}{\textbf{0.58}}} (0.04)& {\color{black}{\textbf{0.44}}} (0.03)& {\color{black}{\textbf{0.46}}} (0.03)& {\color{black}{\textbf{0.45}}} (0.03)\\
 & 5& {\color{black}{\textbf{0.98}}} (0.05)& {\color{black}{\underline{8.16}}} (2.35)& 2.67 (1.76)& {\color{black}{\textbf{0.98}}} (0.04)& {\color{black}{\textbf{0.99}}} (0.04)& {\color{black}{\textbf{0.92}}} (0.02)& {\color{black}{\textbf{0.92}}} (0.07)& {\color{black}{\textbf{0.94}}} (0.03)\\
 & 10& {\color{black}{\textbf{1.55}}} (0.08)& {\color{black}{\underline{13.92}}} (3.39)& {\color{black}{\textbf{4.55}}} (3.09)& {\color{black}\mymk{\textbf{1.44}}} (0.05)& {\color{black}{\textbf{1.73}}} (0.14)& {\color{black}{\textbf{1.58}}} (0.07)& {\color{black}{\textbf{1.56}}} (0.07)& {\color{black}{\textbf{1.56}}} (0.06)\\\hline
\end{tabular} 
}
\end{table*}

  \begin{table*} 
\scalebox{0.6}{
\begin{tabular}{|l|l|c|c|c|c|c|c|c|c|} 
\hline 
Problem & k& PBI& IPBI& HypI& DomRank& MSD& QPBI& APD \\ 
 \hline
 & 2& {\color{black}{\textbf{36.95}}} (6.20)& {\color{black}{\textbf{41.75}}} (12.41)& {\color{black}{\underline{70.82}}} (24.42)& {\color{black}{\textbf{36.93}}} (10.38)& {\color{black}\mymk{\textbf{33.28}}} (12.12)& {\color{black}{\textbf{38.19}}} (11.76)& {\color{black}{\textbf{36.44}}} (4.26)\\
 DTLZ1& 3& {\color{black}{\textbf{34.63}}} (6.66)& {\color{black}{\textbf{43.24}}} (10.63)& {\color{black}{\underline{60.13}}} (21.74)& {\color{black}{\textbf{37.79}}} (15.00)& {\color{black}{\textbf{32.37}}} (10.85)& {\color{black}\mymk{\textbf{30.92}}} (11.70)& {\color{black}{\textbf{34.58}}} (8.18)\\
 & 5& {\color{black}\mymk{\textbf{13.42}}} (8.23)& 36.59 (12.28)& {\color{black}{\underline{48.57}}} (16.87)& 34.74 (10.43)& {\color{black}{\textbf{21.29}}} (13.64)& {\color{black}{\textbf{26.69}}} (5.78)& {\color{black}{\textbf{28.94}}} (10.56)\\
 & 10& {\color{black}{\textbf{31.60}}} (16.09)& {\color{black}{\textbf{36.73}}} (11.10)& {\color{black}{\textbf{39.49}}} (15.73)& {\color{black}{\underline{39.97}}} (14.51)& {\color{black}{\textbf{34.94}}} (7.95)& {\color{black}{\textbf{35.80}}} (14.57)& {\color{black}\mymk{\textbf{29.89}}} (15.99)\\\hline
 & 2& 0.11 (0.01)& 0.08 (0.02)& {\color{black}{\textbf{0.08}}} (0.02)& 0.10 (0.03)& 0.13 (0.05)& 0.09 (0.02)& {\color{black}{\textbf{0.04}}} (0.01)\\
 DTLZ2& 3& 0.15 (0.01)& 0.26 (0.02)& {\color{black}\mymk{\textbf{0.11}}} (0.01)& 0.18 (0.02)& 0.16 (0.03)& 0.15 (0.01)& 0.14 (0.01)\\
 & 5& {\color{black}{\textbf{0.25}}} (0.01)& {\color{black}{\underline{0.45}}} (0.01)& 0.28 (0.02)& 0.36 (0.01)& 0.30 (0.06)& {\color{black}\mymk{\textbf{0.23}}} (0.00)& {\color{black}{\textbf{0.23}}} (0.00)\\
 & 10& 0.83 (0.04)& {\color{black}{\underline{0.85}}} (0.03)& {\color{black}\mymk{\textbf{0.72}}} (0.06)& {\color{black}{\textbf{0.77}}} (0.04)& {\color{black}{\textbf{0.75}}} (0.06)& 0.81 (0.04)& {\color{black}{\textbf{0.74}}} (0.04)\\\hline
 & 2& {\color{black}{\textbf{93.65}}} (14.09)& {\color{black}{\textbf{108.98}}} (20.84)& {\color{black}{\underline{177.05}}} (64.60)& {\color{black}{\textbf{100.48}}} (36.21)& {\color{black}{\textbf{98.52}}} (13.61)& {\color{black}{\textbf{90.55}}} (11.43)& {\color{black}{\textbf{94.19}}} (13.95)\\
 DTLZ3& 3& {\color{black}{\textbf{91.46}}} (28.57)& {\color{black}{\textbf{98.01}}} (27.37)& {\color{black}{\underline{169.47}}} (53.32)& 112.14 (27.98)& {\color{black}{\textbf{78.94}}} (24.08)& {\color{black}\mymk{\textbf{68.90}}} (25.97)& {\color{black}{\textbf{104.18}}} (23.84)\\
 & 5& {\color{black}{\textbf{58.51}}} (39.42)& {\color{black}{\textbf{102.92}}} (24.46)& {\color{black}{\underline{170.63}}} (55.24)& 117.51 (39.97)& {\color{black}{\textbf{75.52}}} (28.84)& {\color{black}{\textbf{101.55}}} (11.68)& {\color{black}\mymk{\textbf{56.92}}} (46.26)\\
 & 10& {\color{black}\mymk{\textbf{82.01}}} (29.42)& {\color{black}{\textbf{108.02}}} (36.65)& {\color{black}{\underline{126.97}}} (55.58)& {\color{black}{\textbf{118.07}}} (23.69)& {\color{black}{\textbf{102.90}}} (27.71)& {\color{black}{\textbf{98.89}}} (11.89)& {\color{black}{\textbf{99.95}}} (46.68)\\\hline
 & 2& {\color{black}{\textbf{0.30}}} (0.06)& {\color{black}{\textbf{0.32}}} (0.05)& {\color{black}{\textbf{0.26}}} (0.08)& {\color{black}{\textbf{0.33}}} (0.05)& {\color{black}{\textbf{0.34}}} (0.03)& {\color{black}{\textbf{0.33}}} (0.07)& {\color{black}\mymk{\textbf{0.22}}} (0.08)\\
 DTLZ4& 3& {\color{black}{\textbf{0.45}}} (0.04)& {\color{black}{\textbf{0.49}}} (0.06)& {\color{black}{\textbf{0.50}}} (0.13)& {\color{black}{\textbf{0.50}}} (0.04)& {\color{black}{\textbf{0.41}}} (0.09)& {\color{black}{\textbf{0.42}}} (0.03)& {\color{black}\mymk{\textbf{0.41}}} (0.07)\\
 & 5& {\color{black}\mymk{\textbf{0.55}}} (0.02)& 0.73 (0.09)& {\color{black}{\underline{0.82}}} (0.09)& {\color{black}{\textbf{0.63}}} (0.05)& 0.71 (0.08)& {\color{black}{\textbf{0.58}}} (0.04)& {\color{black}{\textbf{0.56}}} (0.02)\\
 & 10& {\color{black}\mymk{\textbf{0.72}}} (0.04)& 0.88 (0.02)& {\color{black}{\underline{0.91}}} (0.03)& 0.78 (0.03)& {\color{black}{\textbf{0.76}}} (0.03)& {\color{black}{\textbf{0.72}}} (0.03)& 0.78 (0.04)\\\hline
 & 2& 0.12 (0.01)& {\color{black}{\textbf{0.08}}} (0.01)& {\color{black}{\textbf{0.07}}} (0.01)& 0.09 (0.02)& 0.14 (0.06)& 0.09 (0.01)& {\color{black}\mymk{\textbf{0.04}}} (0.01)\\
 DTLZ5& 3& {\color{black}{\textbf{0.05}}} (0.01)& 0.14 (0.02)& {\color{black}{\textbf{0.05}}} (0.01)& 0.09 (0.02)& 0.08 (0.02)& {\color{black}{\textbf{0.04}}} (0.01)& {\color{black}{\textbf{0.06}}} (0.01)\\
 & 5& 0.08 (0.02)& 0.17 (0.02)& 0.07 (0.01)& 0.08 (0.02)& 0.06 (0.02)& {\color{black}{\textbf{0.03}}} (0.01)& {\color{black}\mymk{\textbf{0.02}}} (0.00)\\
 & 10& 0.13 (0.03)& 0.15 (0.02)& 0.10 (0.02)& 0.09 (0.02)& {\color{black}{\textbf{0.08}}} (0.02)& 0.09 (0.01)& {\color{black}\mymk{\textbf{0.05}}} (0.01)\\\hline
 & 2& {\color{black}{\textbf{0.32}}} (0.10)& 1.48 (0.24)& {\color{black}{\textbf{0.48}}} (0.22)& 0.73 (0.51)& {\color{black}{\textbf{0.36}}} (0.08)& 0.63 (0.28)& {\color{black}{\textbf{0.41}}} (0.18)\\
 DTLZ6& 3& {\color{black}\mymk{\textbf{0.24}}} (0.04)& 1.53 (0.44)& 1.50 (0.62)& 2.69 (0.66)& 1.68 (1.71)& {\color{black}{\textbf{0.57}}} (0.15)& {\color{black}{\textbf{0.24}}} (0.05)\\
 & 5& {\color{black}\mymk{\textbf{0.31}}} (0.04)& 2.33 (0.53)& 2.48 (0.93)& 2.52 (0.76)& 3.60 (0.51)& 2.24 (0.38)& {\color{black}{\textbf{0.34}}} (0.05)\\
 & 10& {\color{black}\mymk{\textbf{0.55}}} (0.08)& 3.55 (0.47)& 3.20 (0.45)& 3.26 (0.51)& {\color{black}{\underline{3.61}}} (0.45)& 2.43 (0.50)& {\color{black}{\textbf{1.01}}} (0.50)\\\hline
 & 2& {\color{black}{\textbf{0.24}}} (0.02)& {\color{black}{\textbf{0.37}}} (0.00)& {\color{black}\mymk{\textbf{0.08}}} (0.02)& {\color{black}{\textbf{0.41}}} (0.11)& {\color{black}{\textbf{0.32}}} (0.21)& 0.73 (0.13)& {\color{black}{\textbf{0.23}}} (0.09)\\
 DTLZ7& 3& 0.81 (0.13)& 0.85 (0.07)& {\color{black}\mymk{\textbf{0.22}}} (0.12)& {\color{black}{\textbf{0.73}}} (0.14)& {\color{black}{\textbf{0.63}}} (0.32)& 1.25 (0.33)& {\color{black}{\textbf{0.67}}} (0.19)\\
 & 5& {\color{black}{\textbf{2.20}}} (0.36)& 5.15 (2.82)& {\color{black}\mymk{\textbf{0.76}}} (0.13)& {\color{black}{\textbf{2.20}}} (0.62)& {\color{black}{\textbf{1.95}}} (0.93)& 2.93 (0.23)& {\color{black}{\textbf{2.03}}} (0.40)\\
 & 10& {\color{black}{\textbf{3.84}}} (0.49)& 13.56 (3.45)& 6.86 (5.17)& 7.89 (4.05)& 11.42 (4.56)& 8.28 (2.66)& 6.03 (2.87) \\ \hline
\end{tabular} 
}
\caption{\label{tab:IGD_DTLZ}Mean IGD values and standard deviation (in parentheses) for DTLZ problems. The values statistically similar to the best one are in bold, the best value is encircled the worst value is underlined}
\end{table*}

\subsection{Sensitivity towards number of objectives}
\begin{figure*}
    \centering
    \subfloat[]{\includegraphics[width=0.4\textwidth]{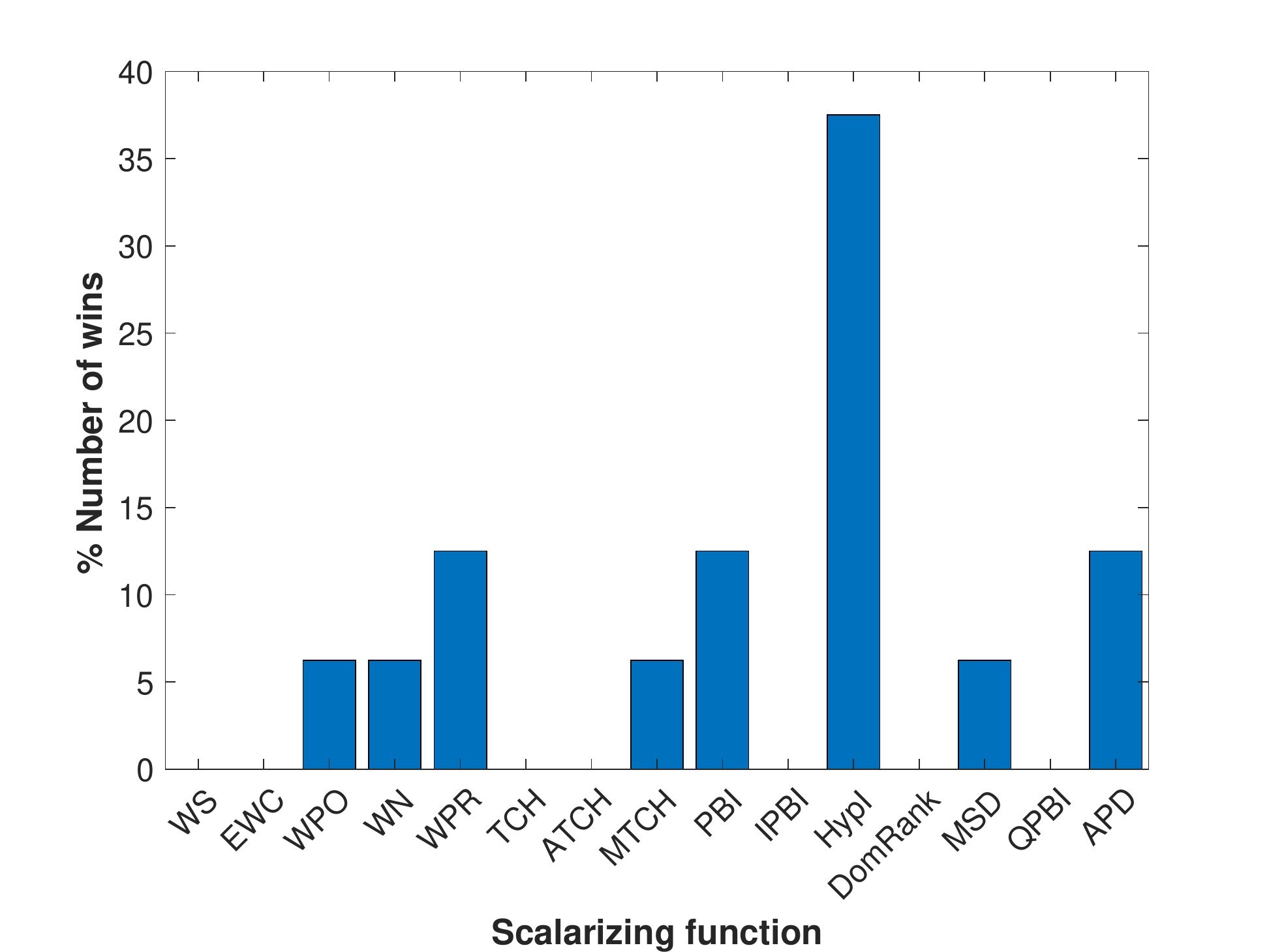}}
     \subfloat[]{\includegraphics[width=0.4\textwidth]{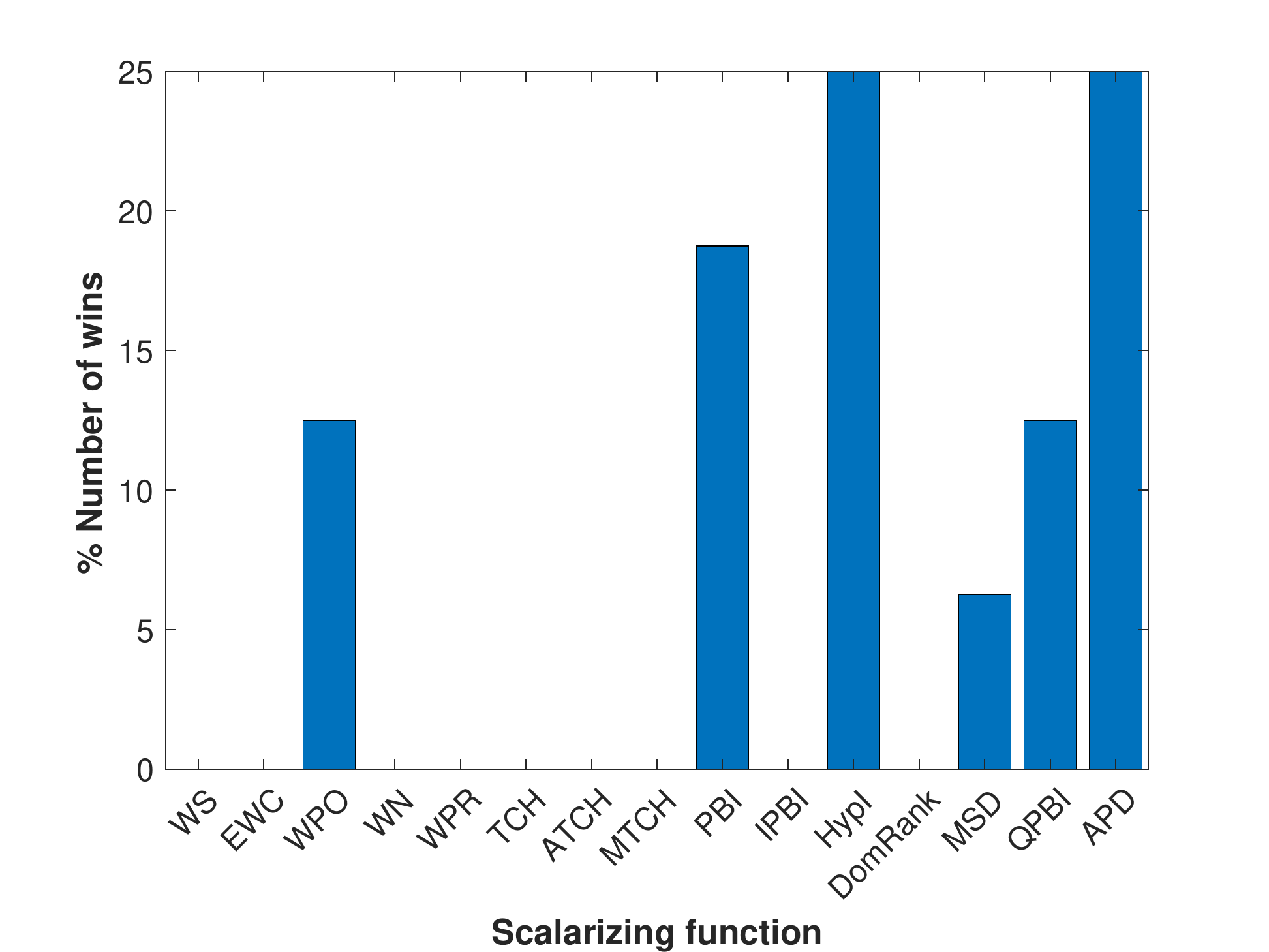}} \\
     \subfloat[]{\includegraphics[width=0.4\textwidth]{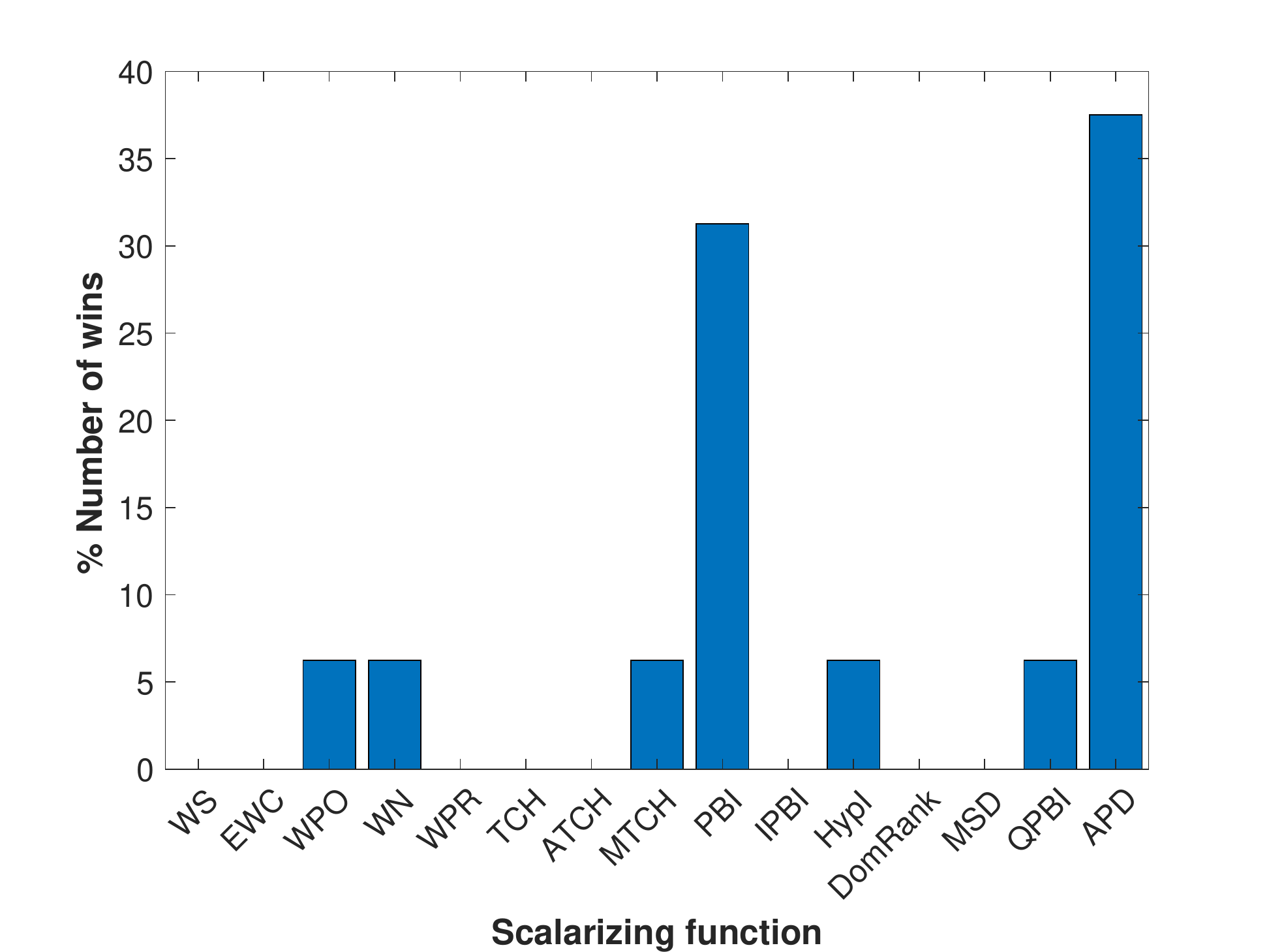}}
     \subfloat[]{\includegraphics[width=0.4\textwidth]{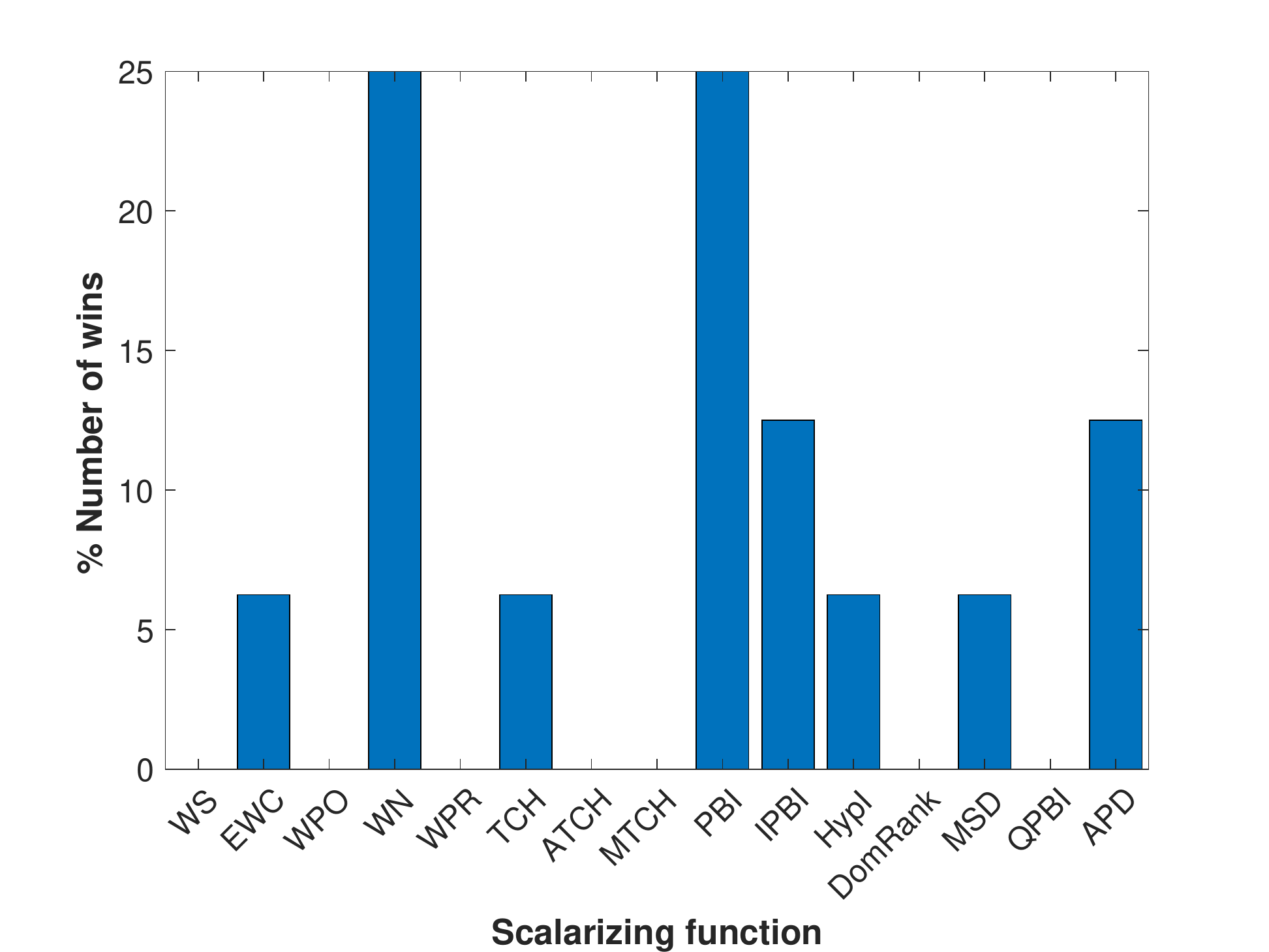}}
     \caption{\label{fig:bar_graph_objectives}Percentage number of wins of different scalarizing functions with respect to a) two b) three c) five and d) 10 objectives.}
\end{figure*}

When using scalarizing functions for building surrogates on them, the number of objectives can play a crucial role in their performance. Before using the surrogate built on the scalarizing function, we can analyze and visualize the fitness landscape with respect to the accuracy and uncertainty of approximations (or predictions). We start by showing the bar graph of the number of wins with respect to 2, 3, 5 and 10 objectives in Figure \ref{fig:bar_graph_objectives}.

\begin{figure*}
    \centering
    \includegraphics[scale=0.2]{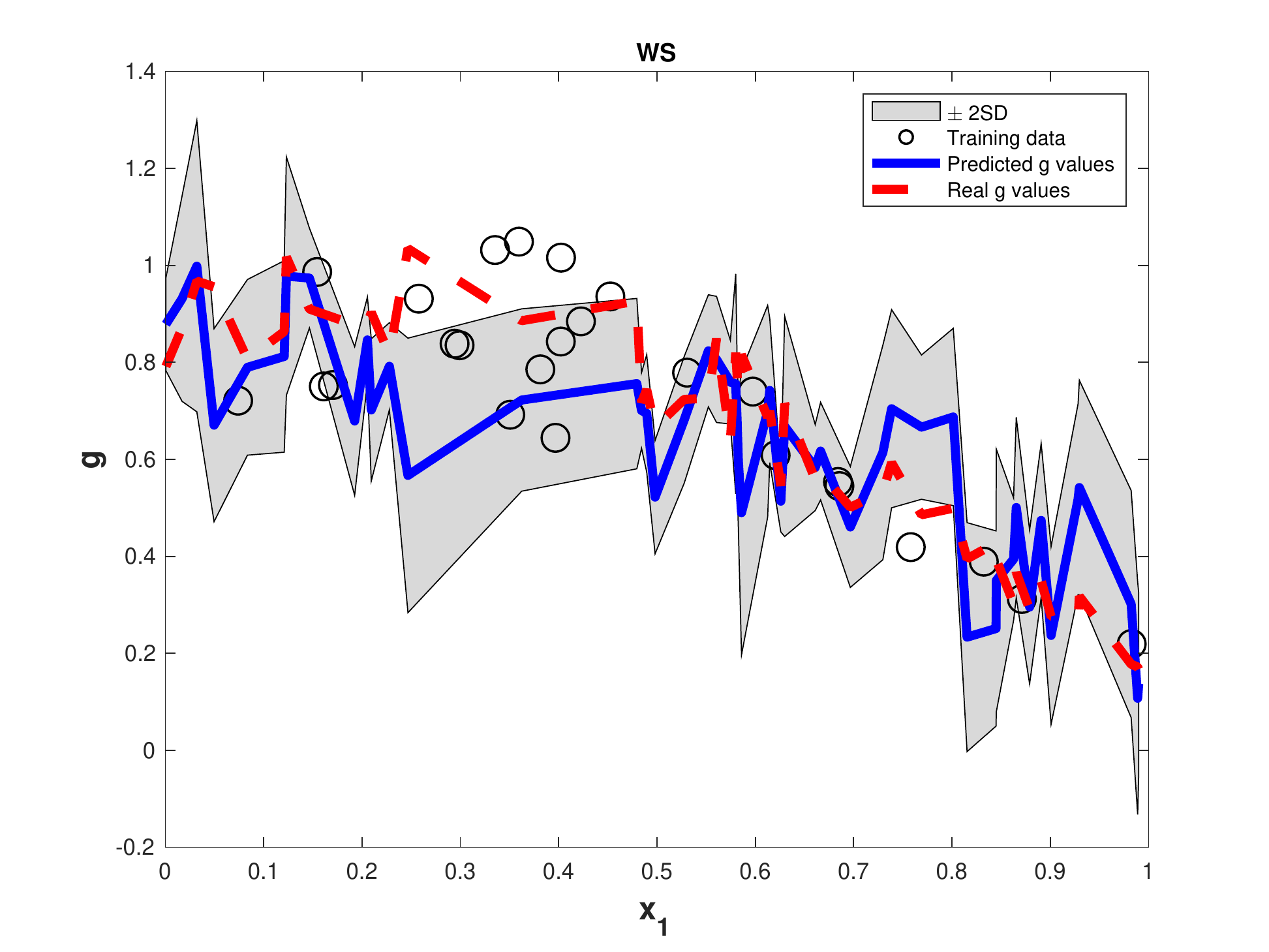}
    \includegraphics[scale=0.2]{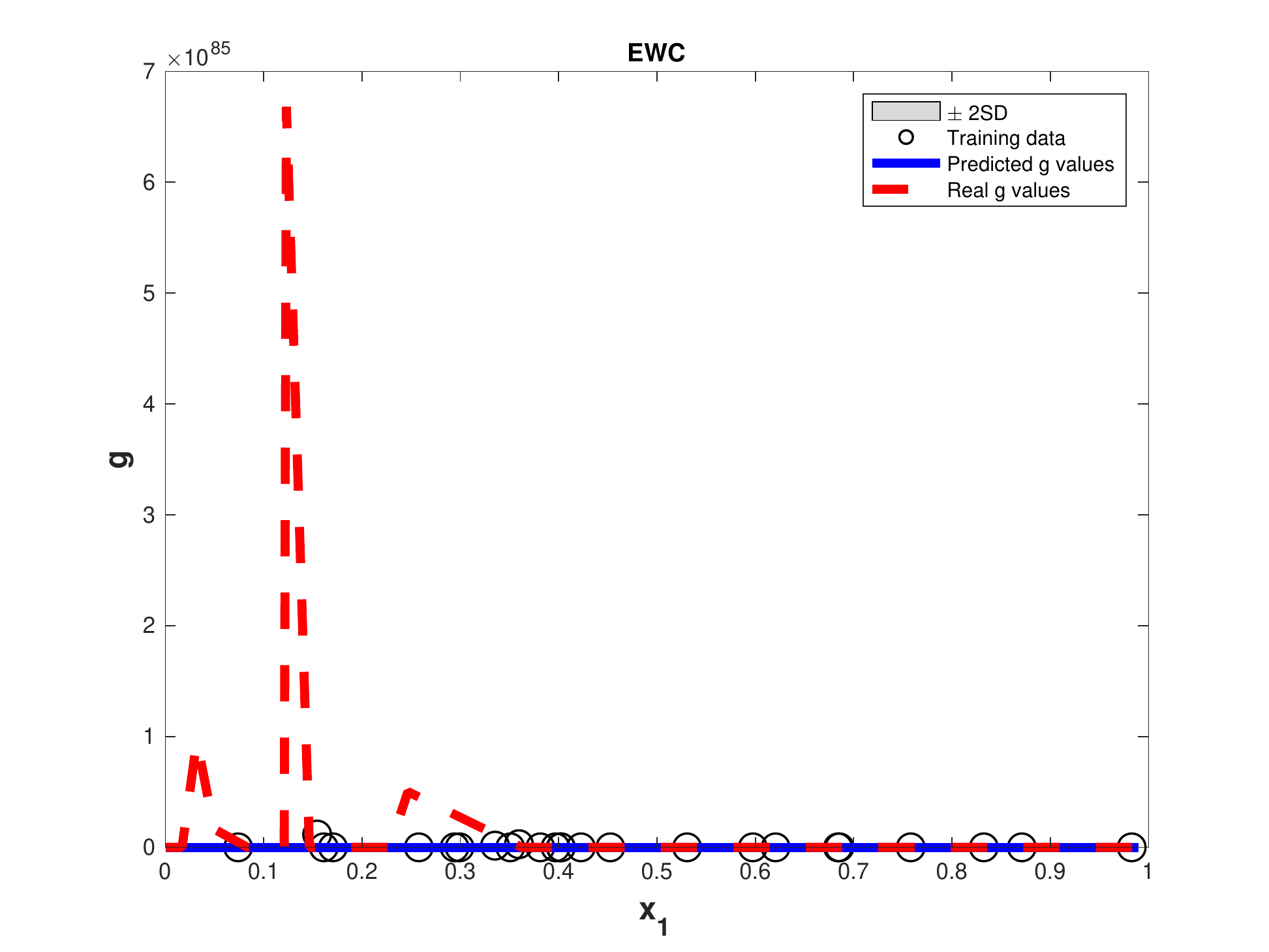}
    \includegraphics[scale=0.2]{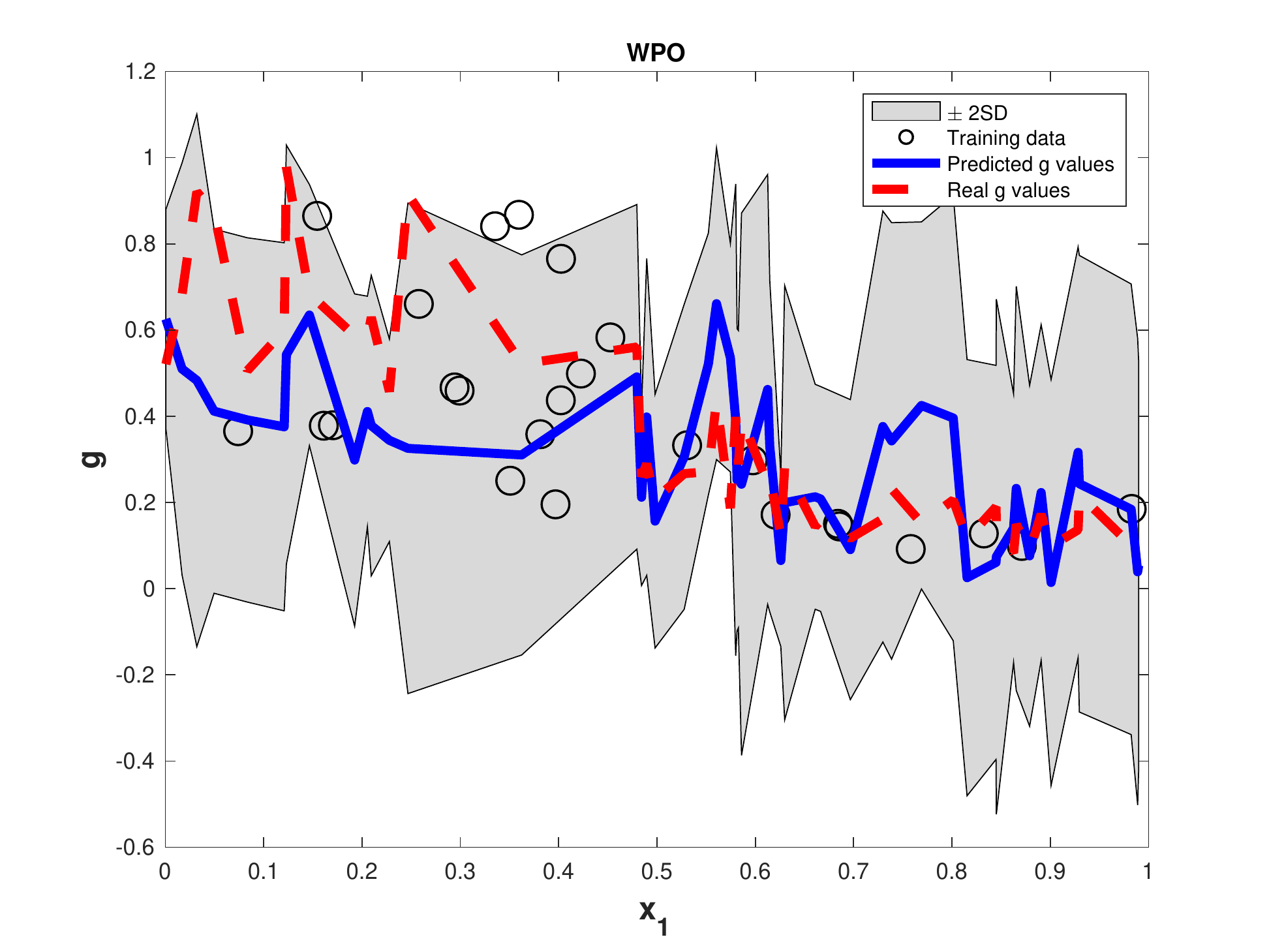}
    \includegraphics[scale=0.2]{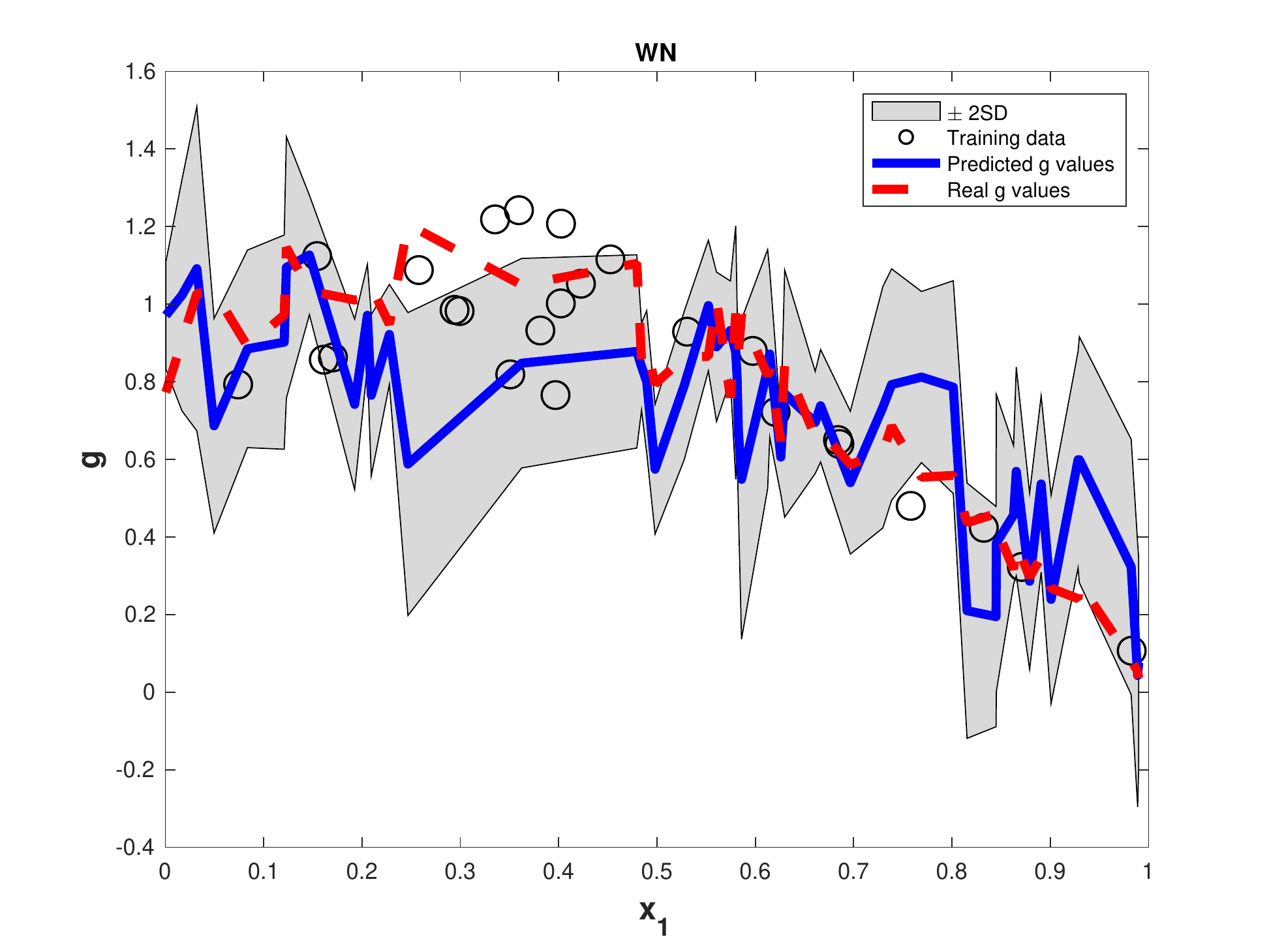}
    \includegraphics[scale=0.2]{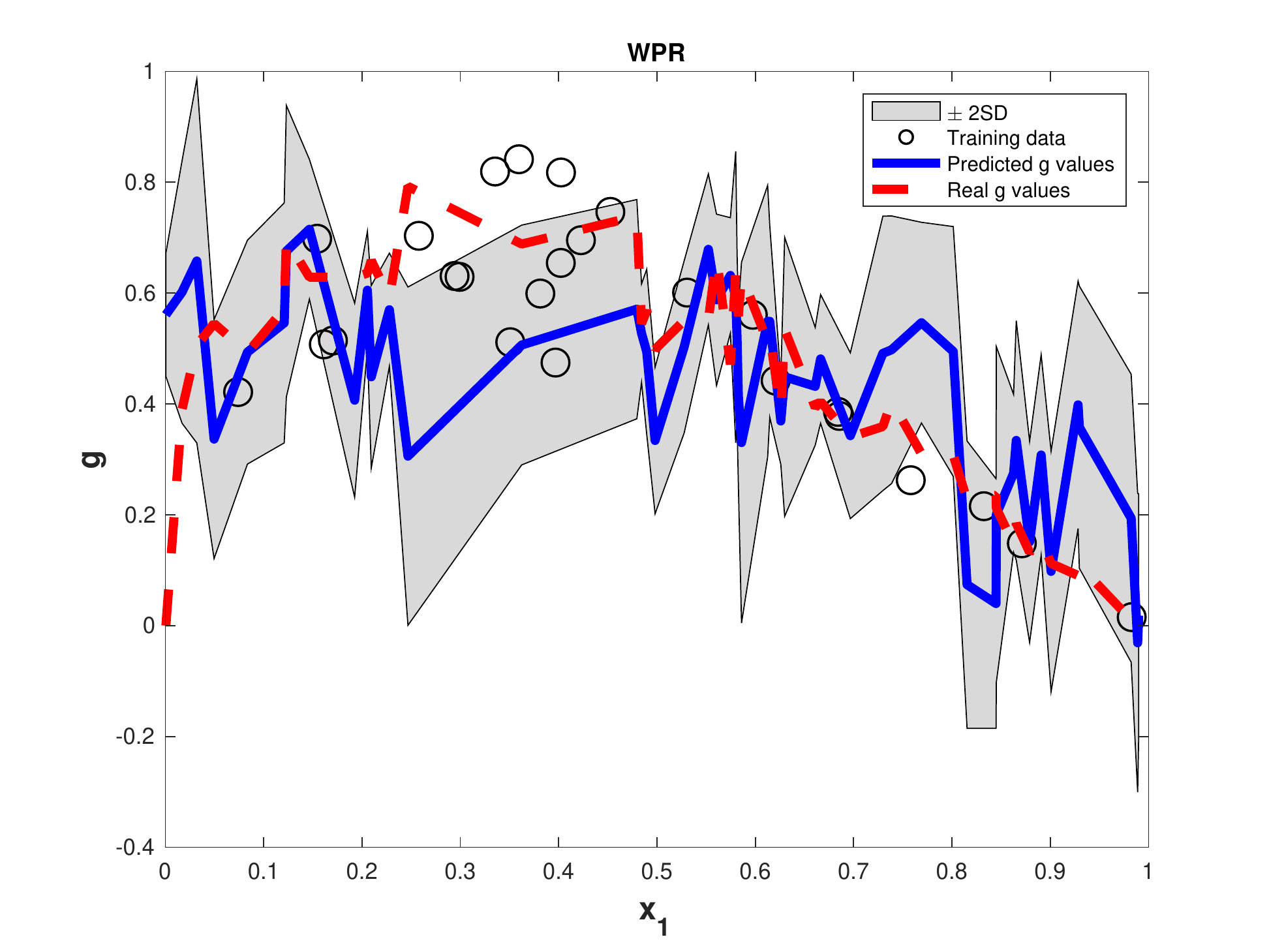}
    \includegraphics[scale=0.2]{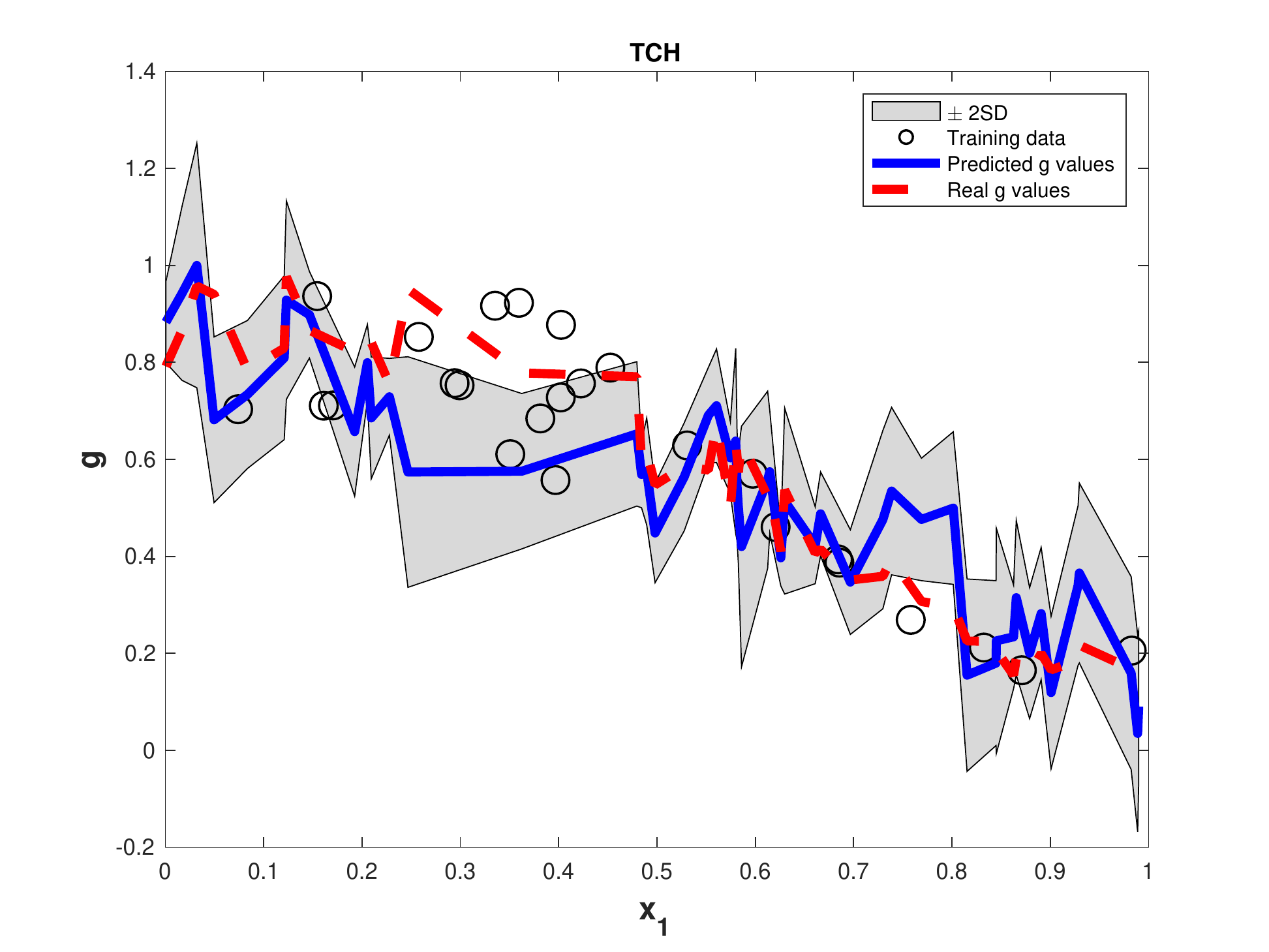}
    \includegraphics[scale=0.2]{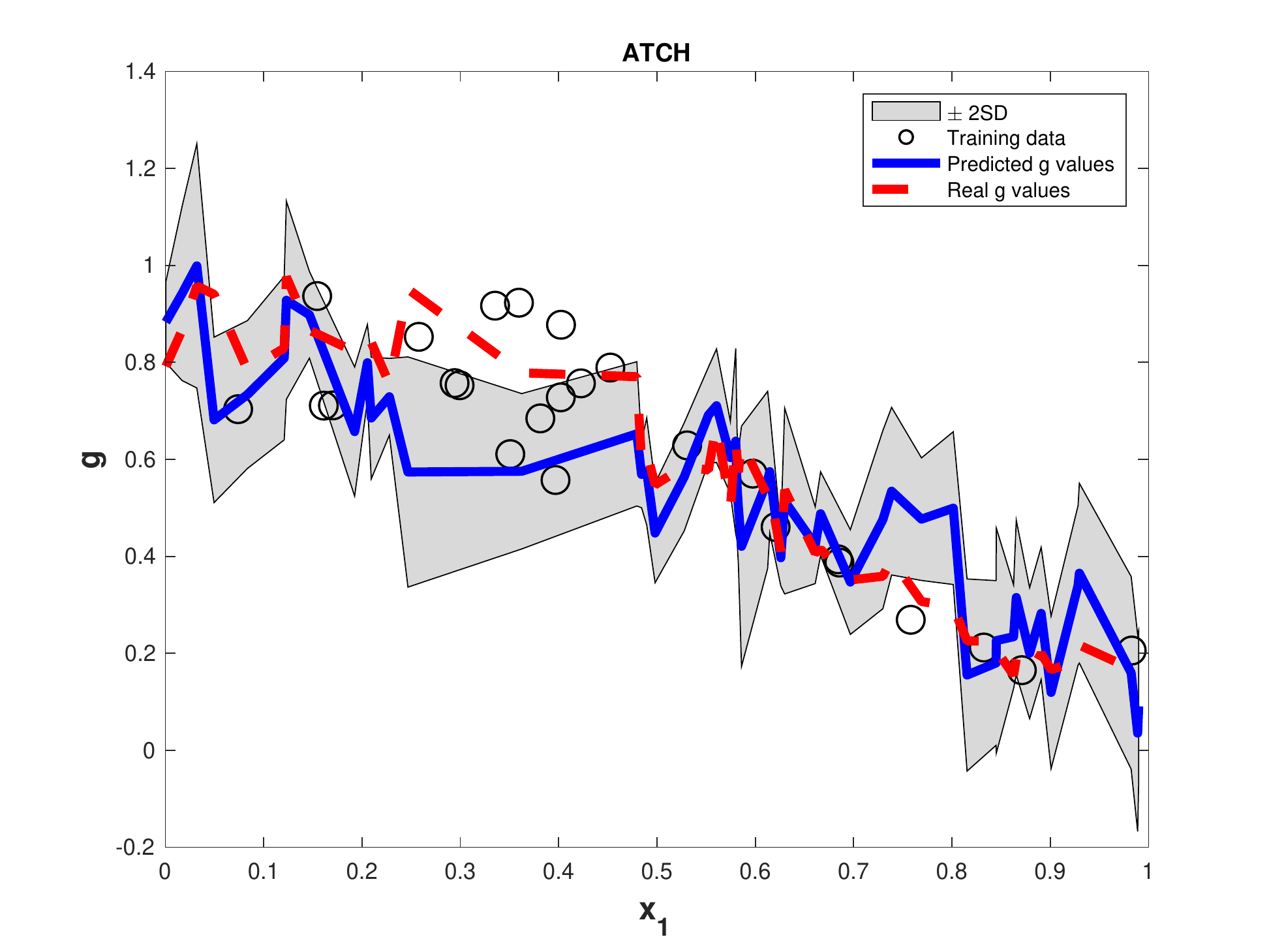}
    \includegraphics[scale=0.2]{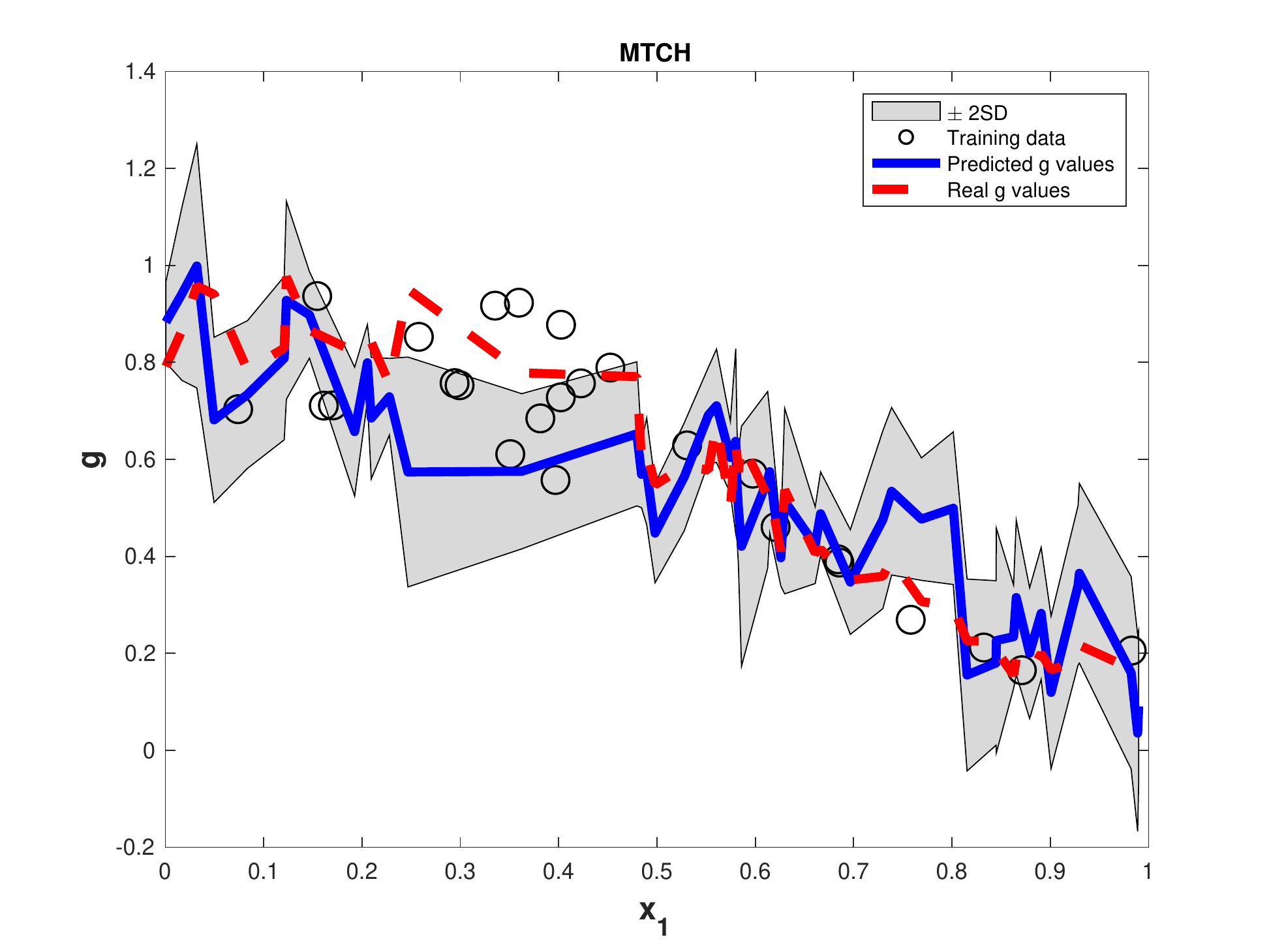}
    \includegraphics[scale=0.2]{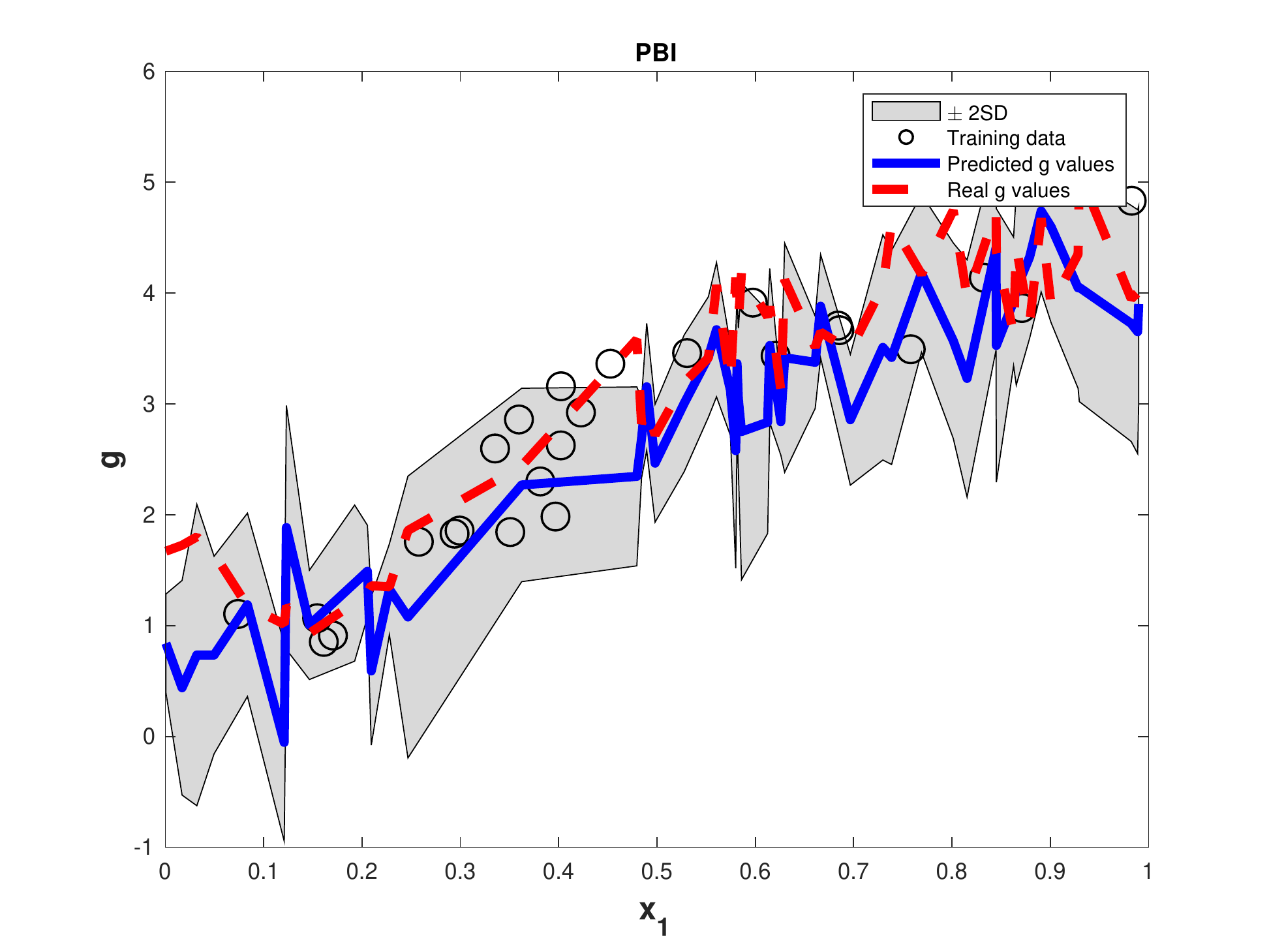}
    \includegraphics[scale=0.2]{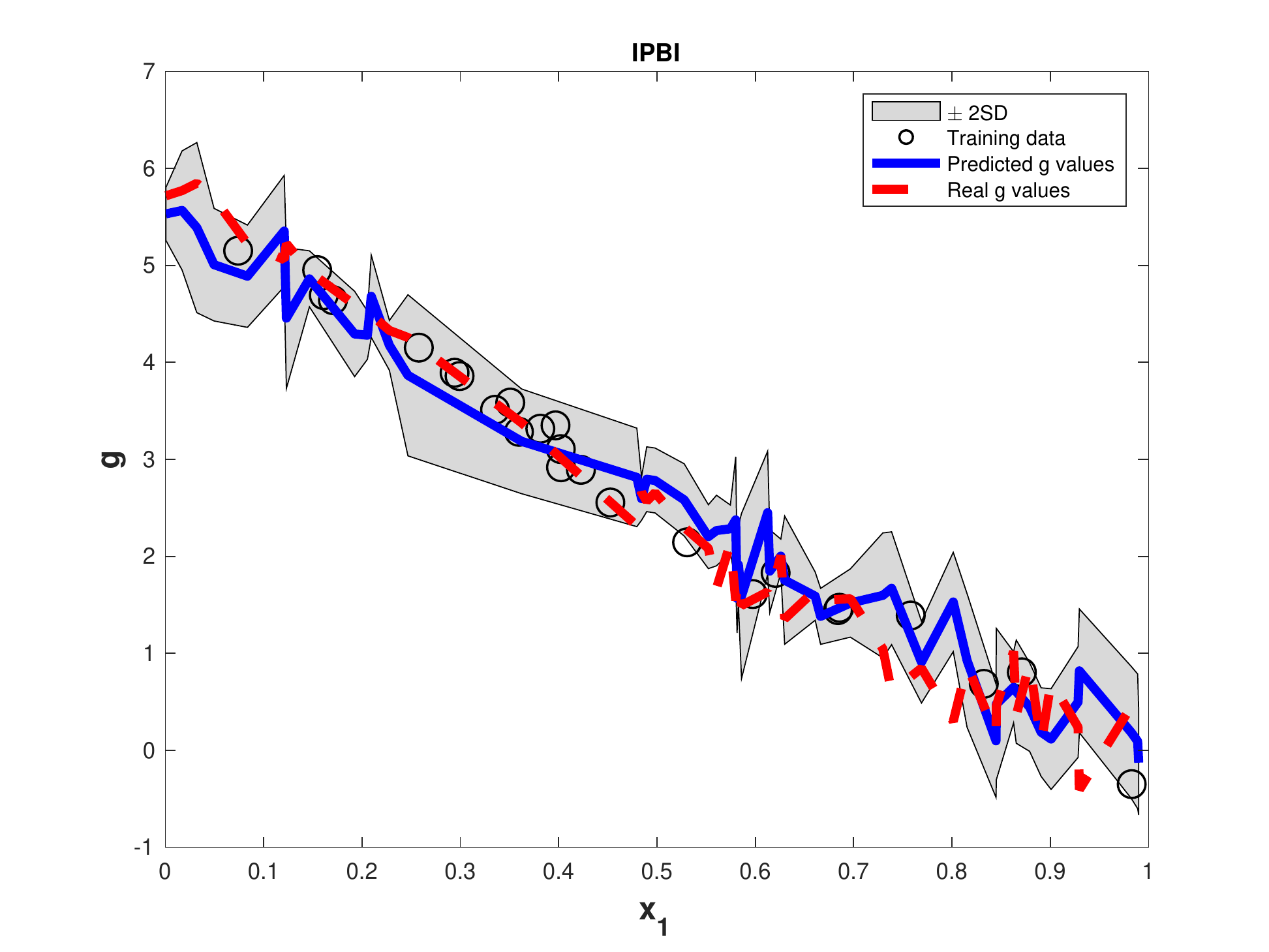}
    \includegraphics[scale=0.2]{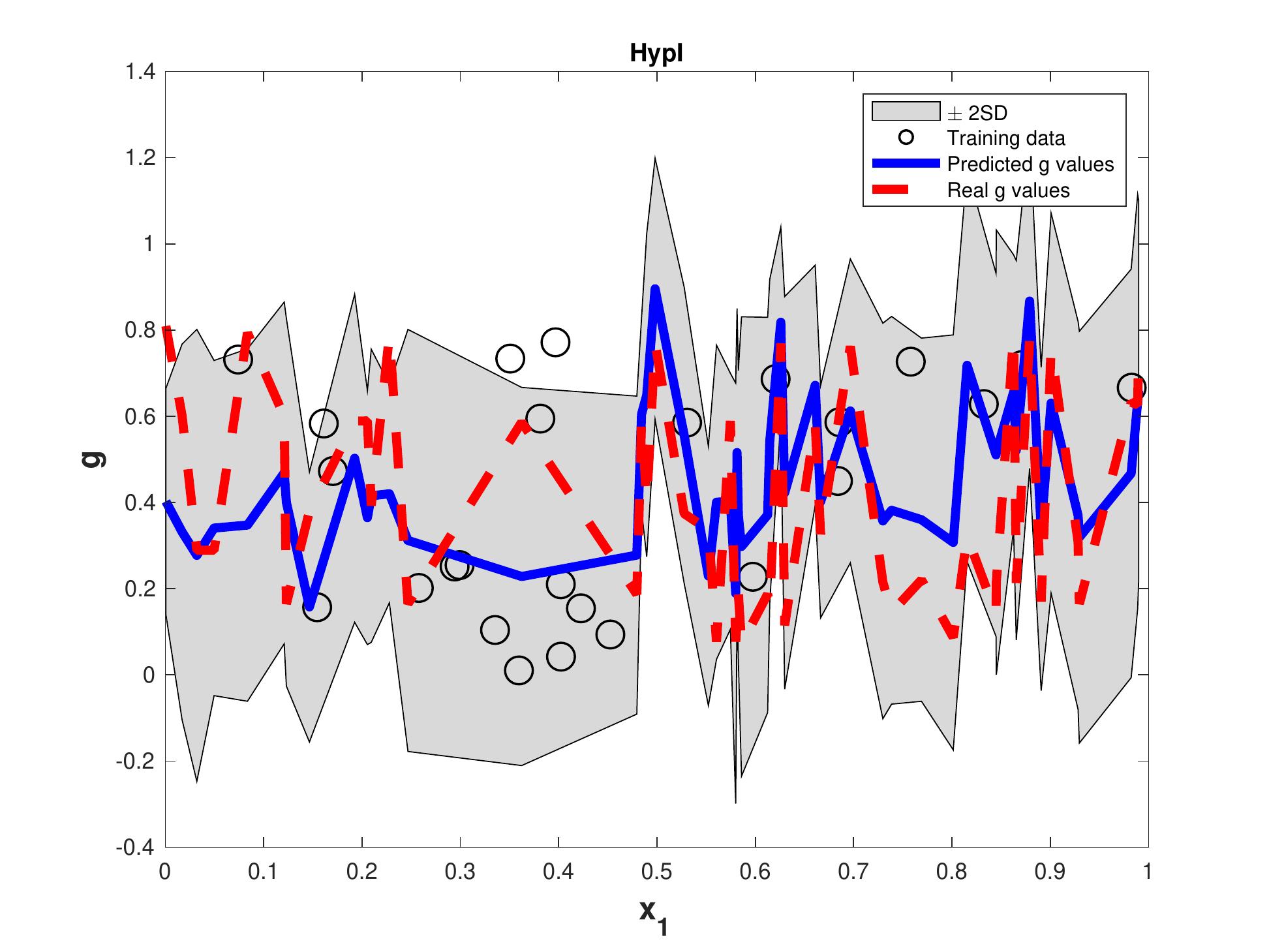}
    \includegraphics[scale=0.2]{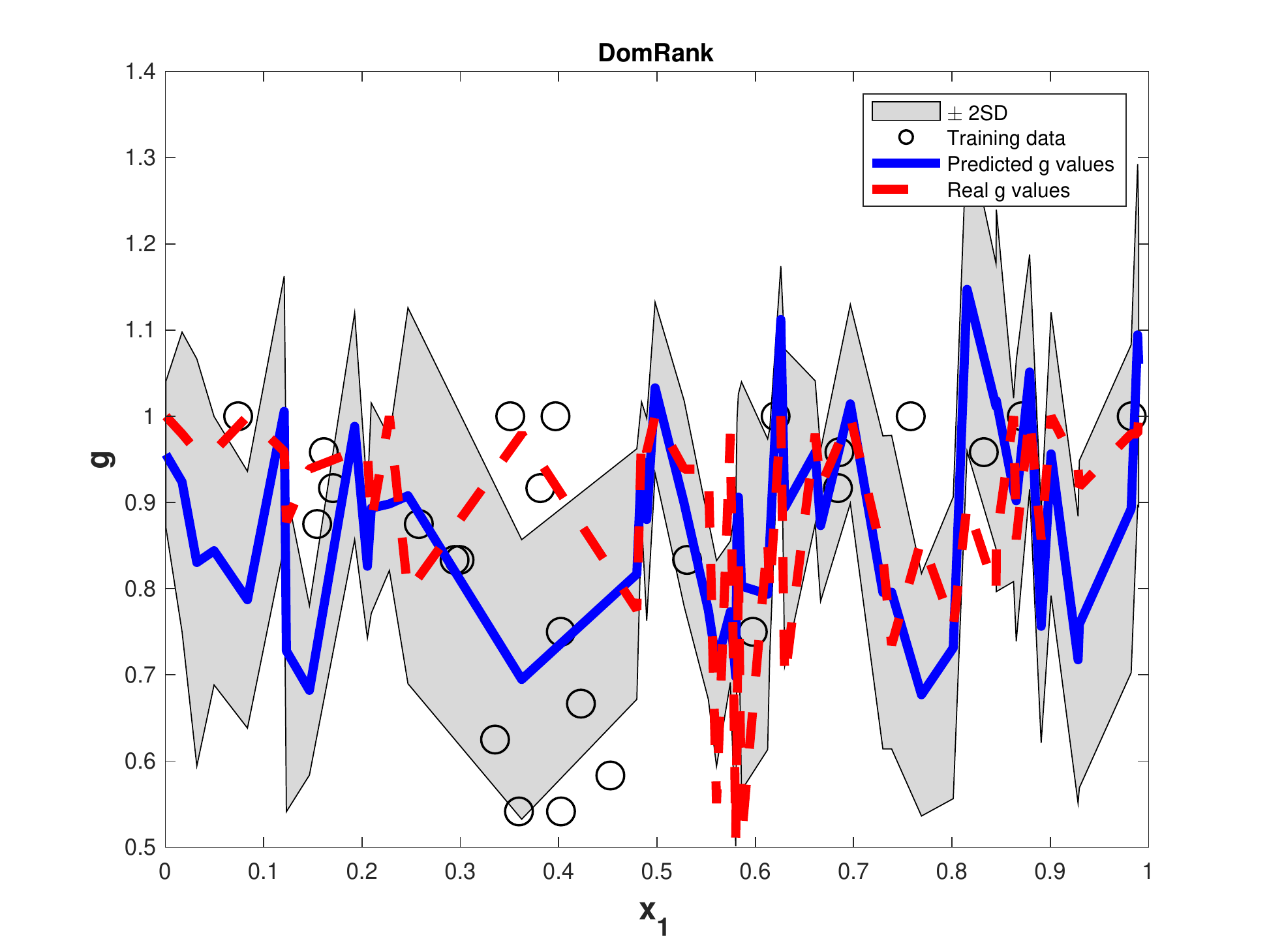}
    \includegraphics[scale=0.2]{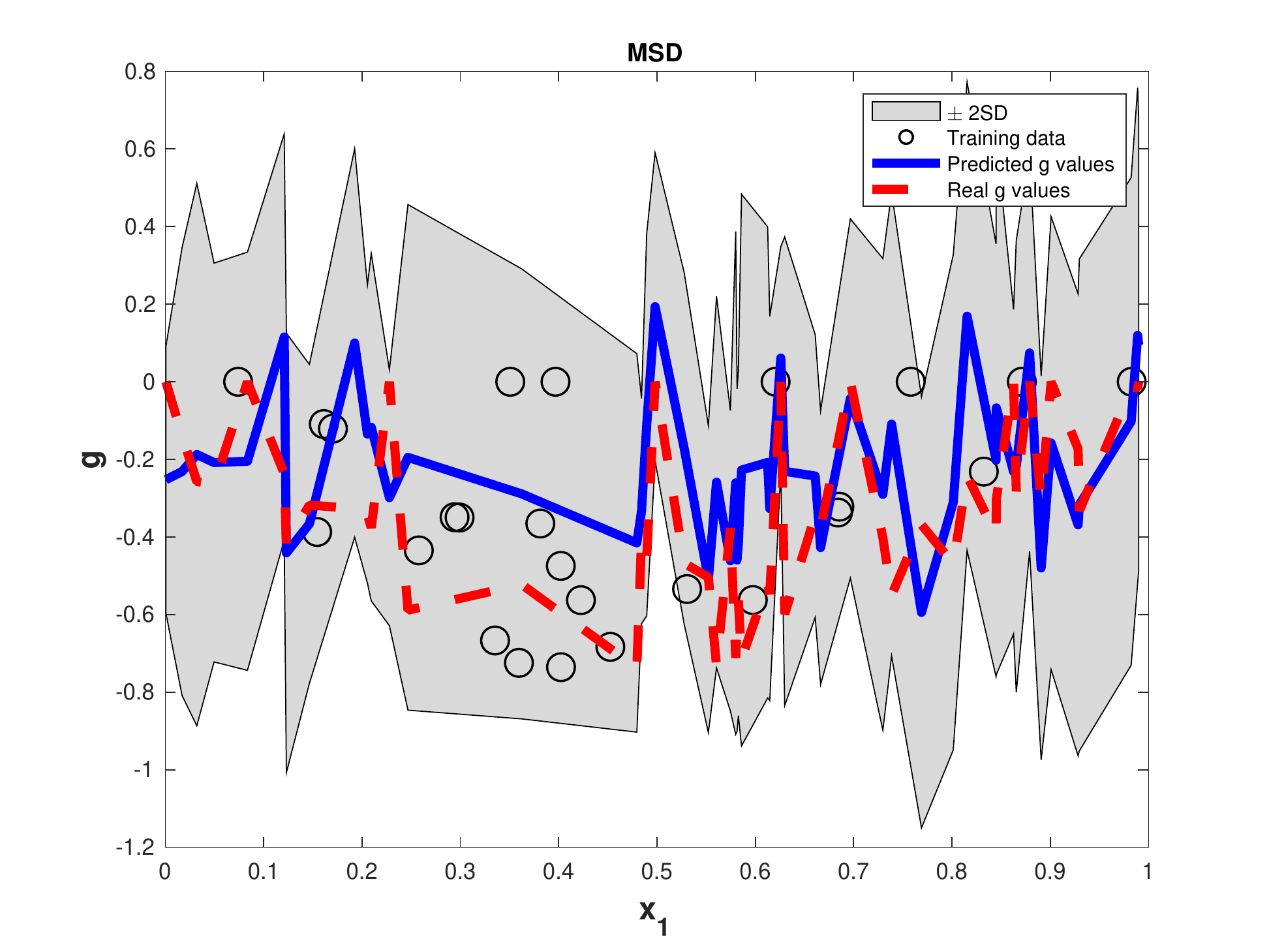}
    \includegraphics[scale=0.2]{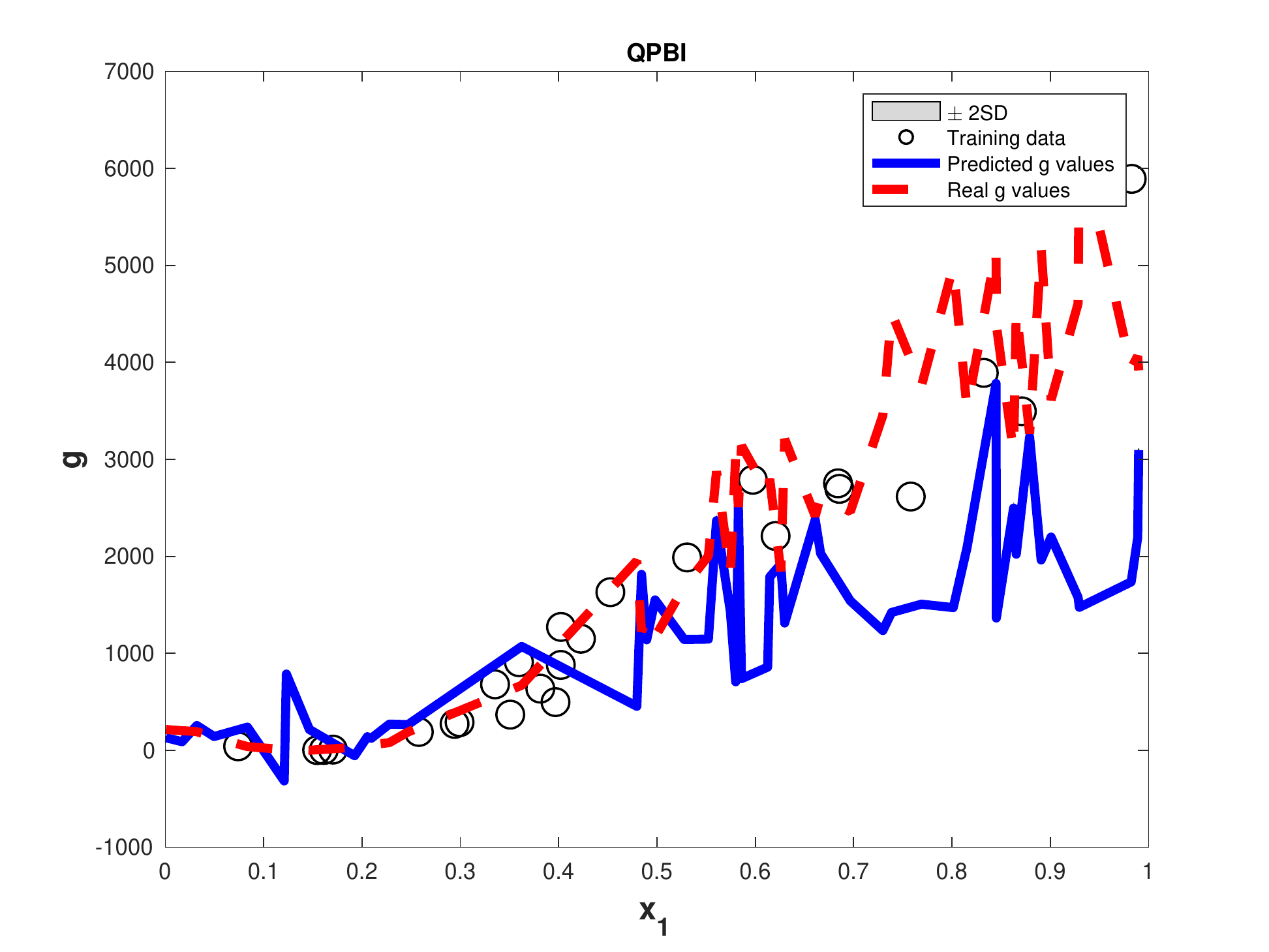}
    \includegraphics[scale=0.2]{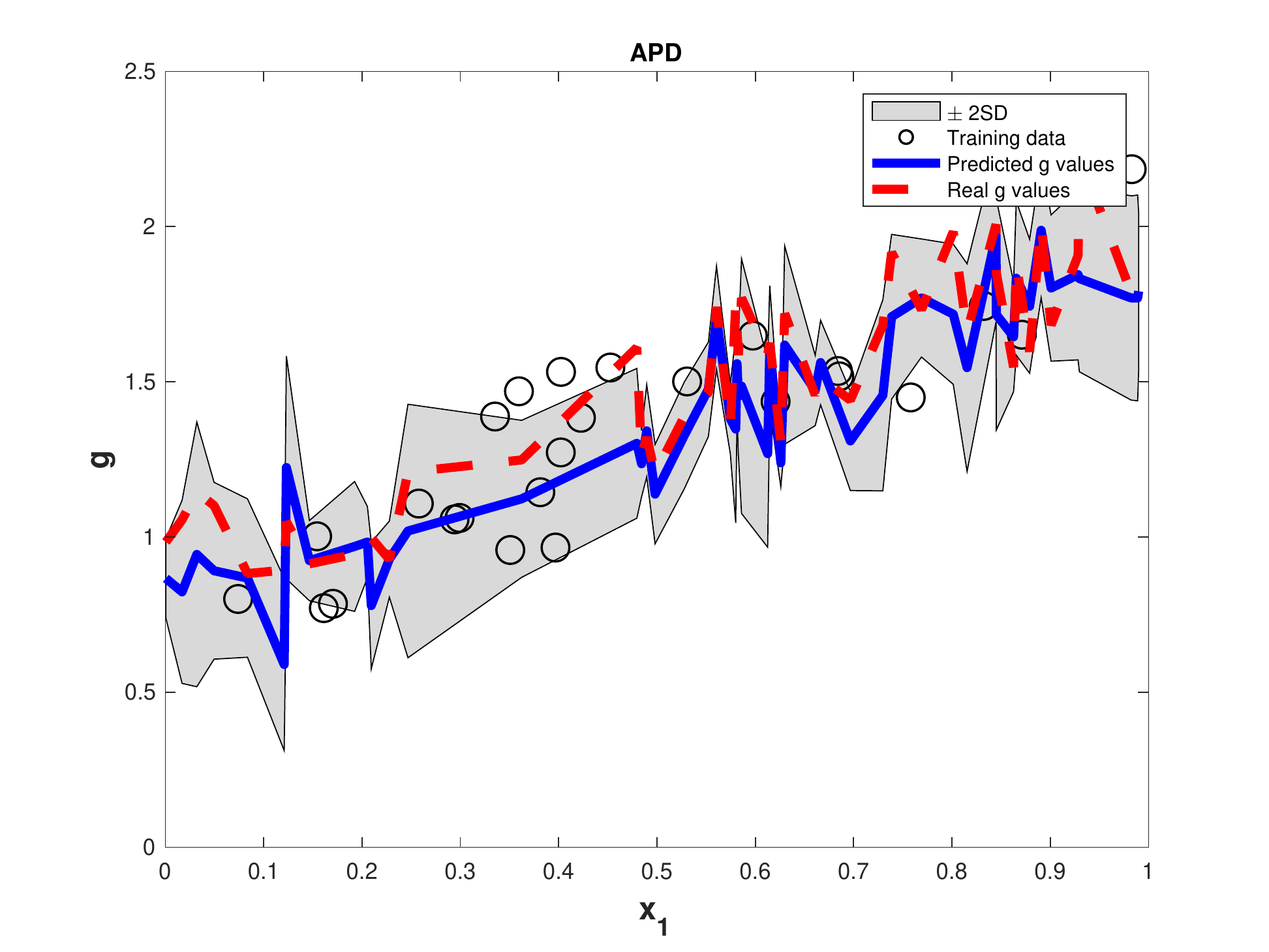}
    \caption{\label{fig:fitness_landscape_DTLZ2_2}Approximated scalarizing function values (notated by $g$) with one decision variable value for DTLZ2 two objectives, $\pm $2SD represents the approximated $g$ values with $\pm$ 2 standard deviations of the approximated values}
\end{figure*}

\begin{figure*}
    \centering
    \includegraphics[scale=0.2]{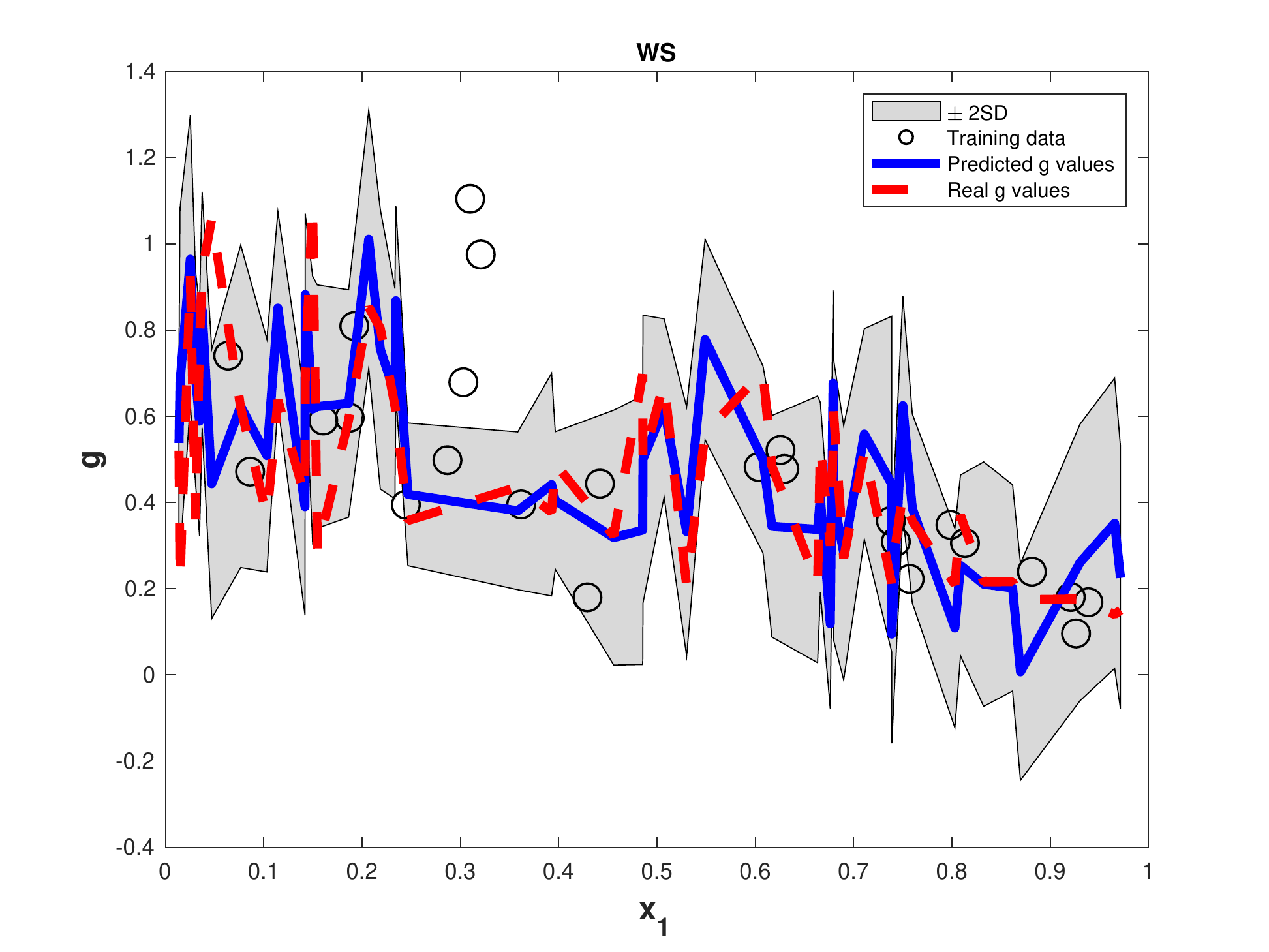}
    \includegraphics[scale=0.2]{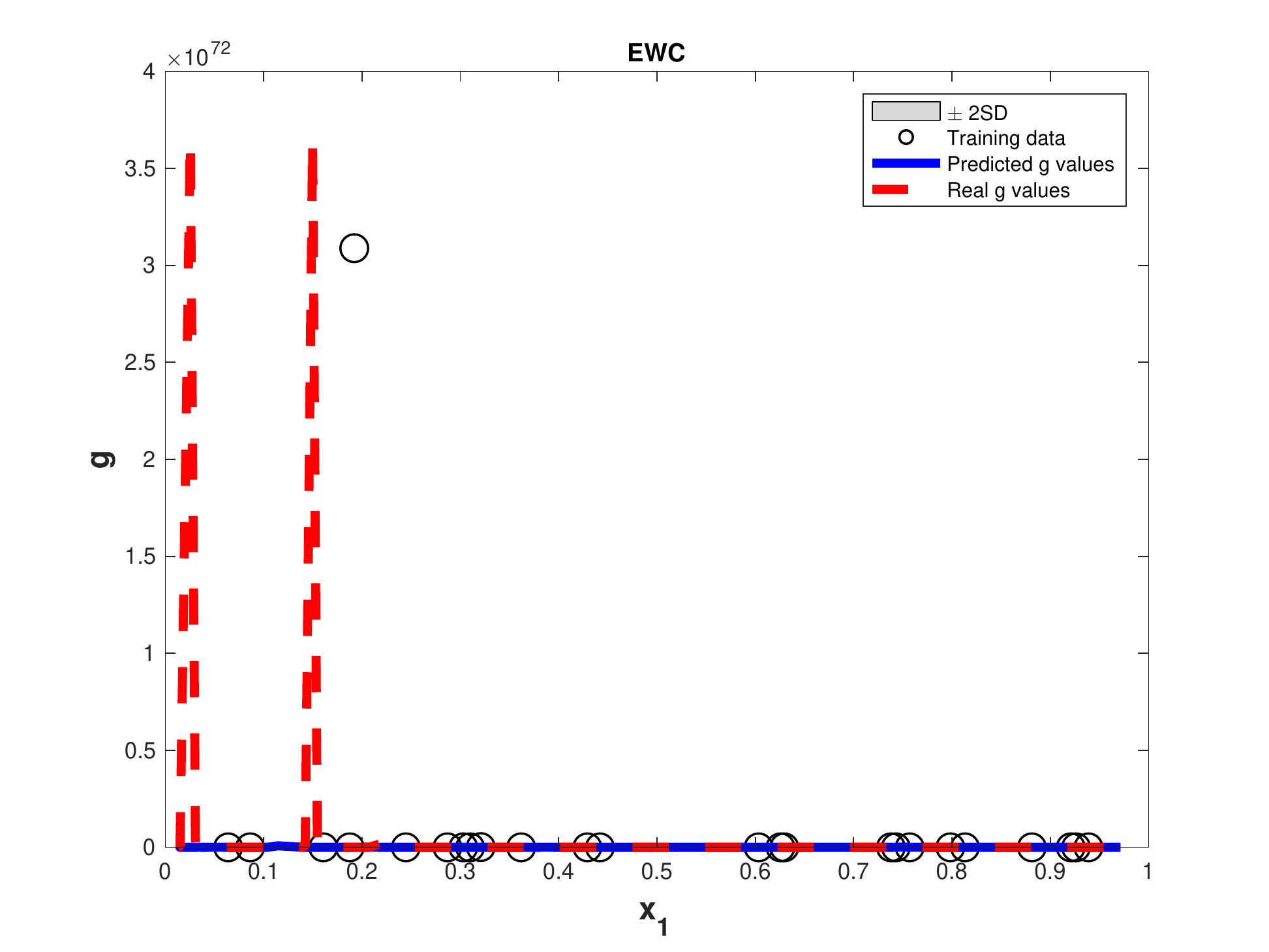}
    \includegraphics[scale=0.2]{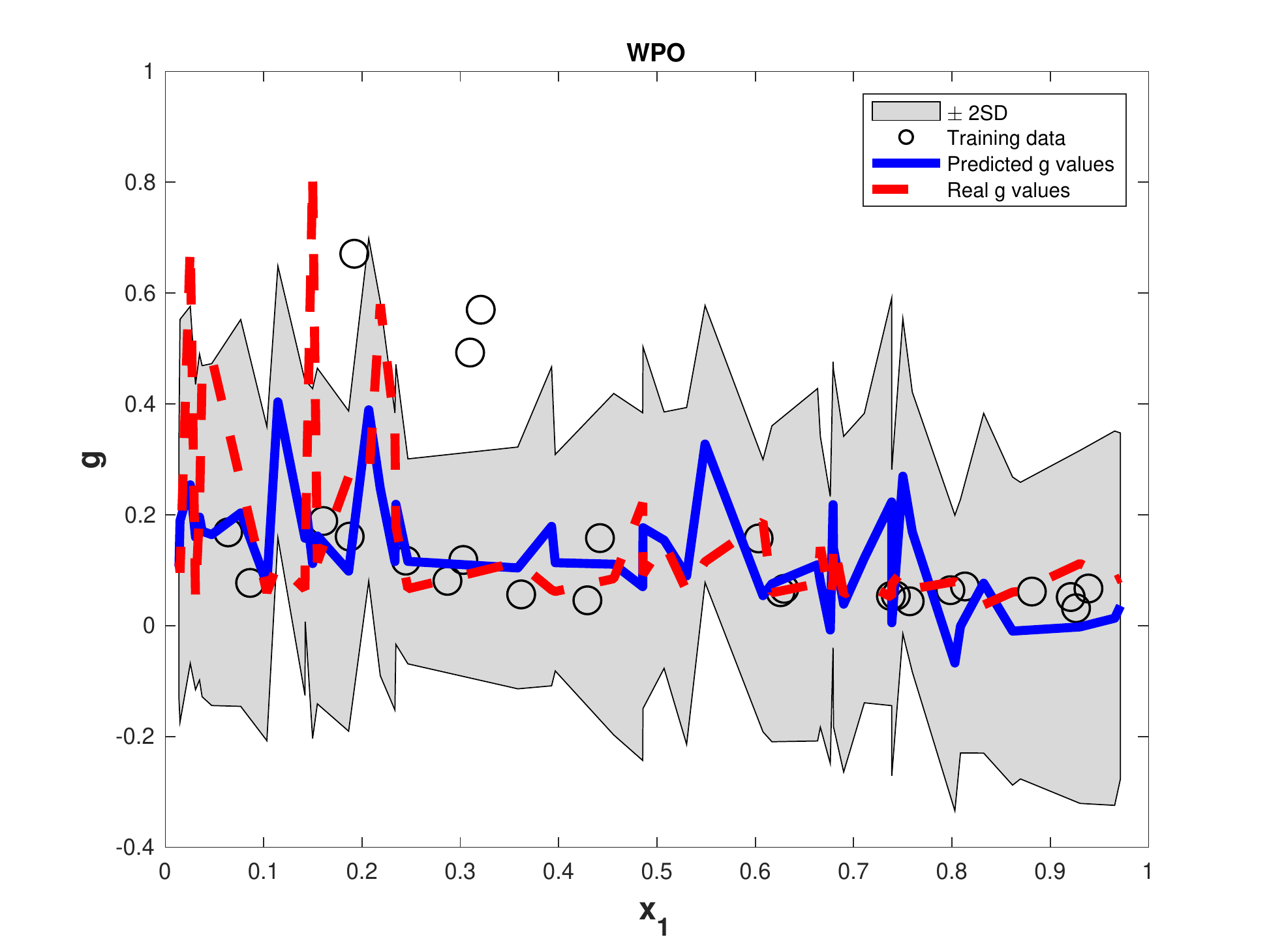}
    \includegraphics[scale=0.2]{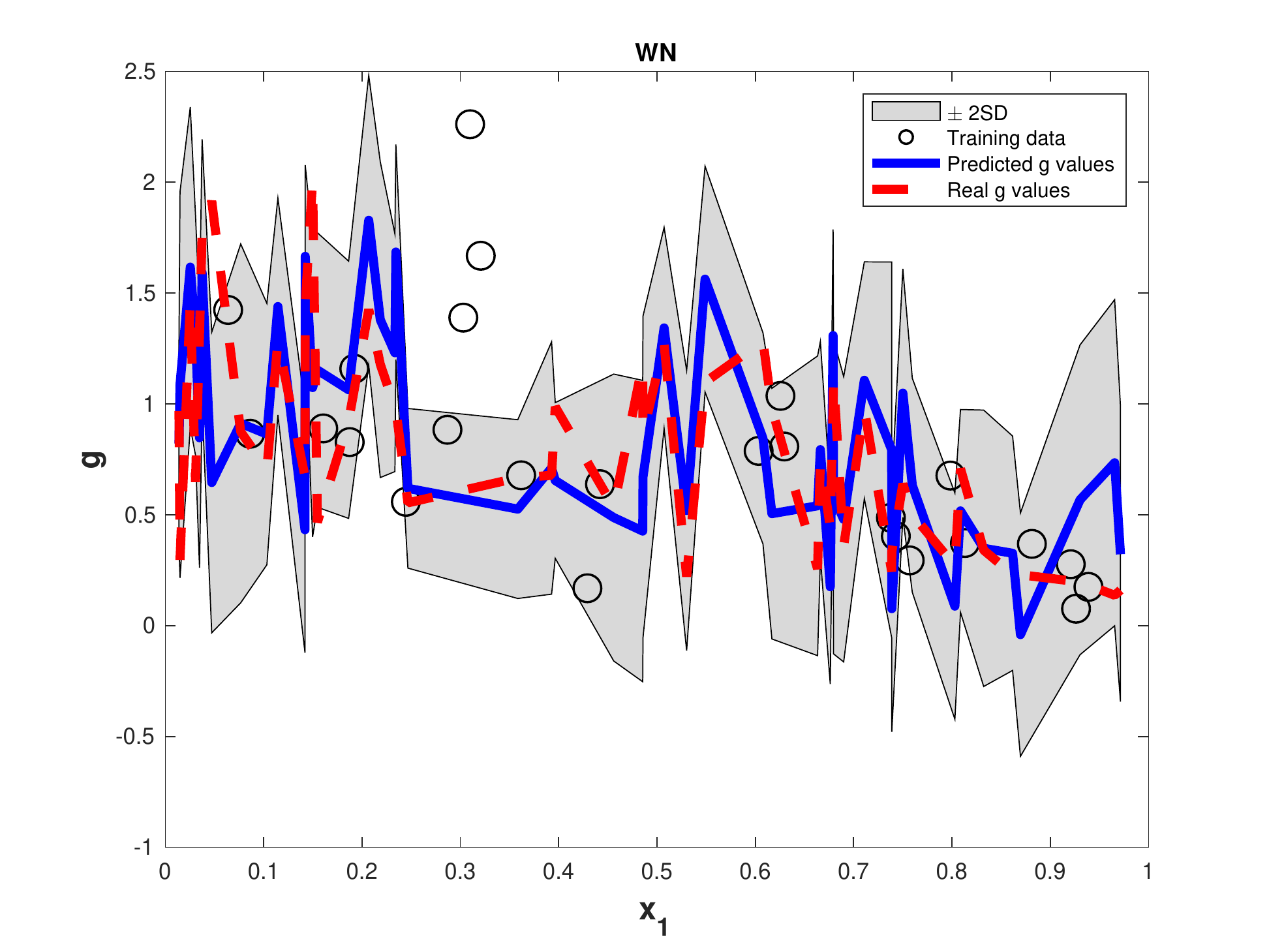}
    \includegraphics[scale=0.2]{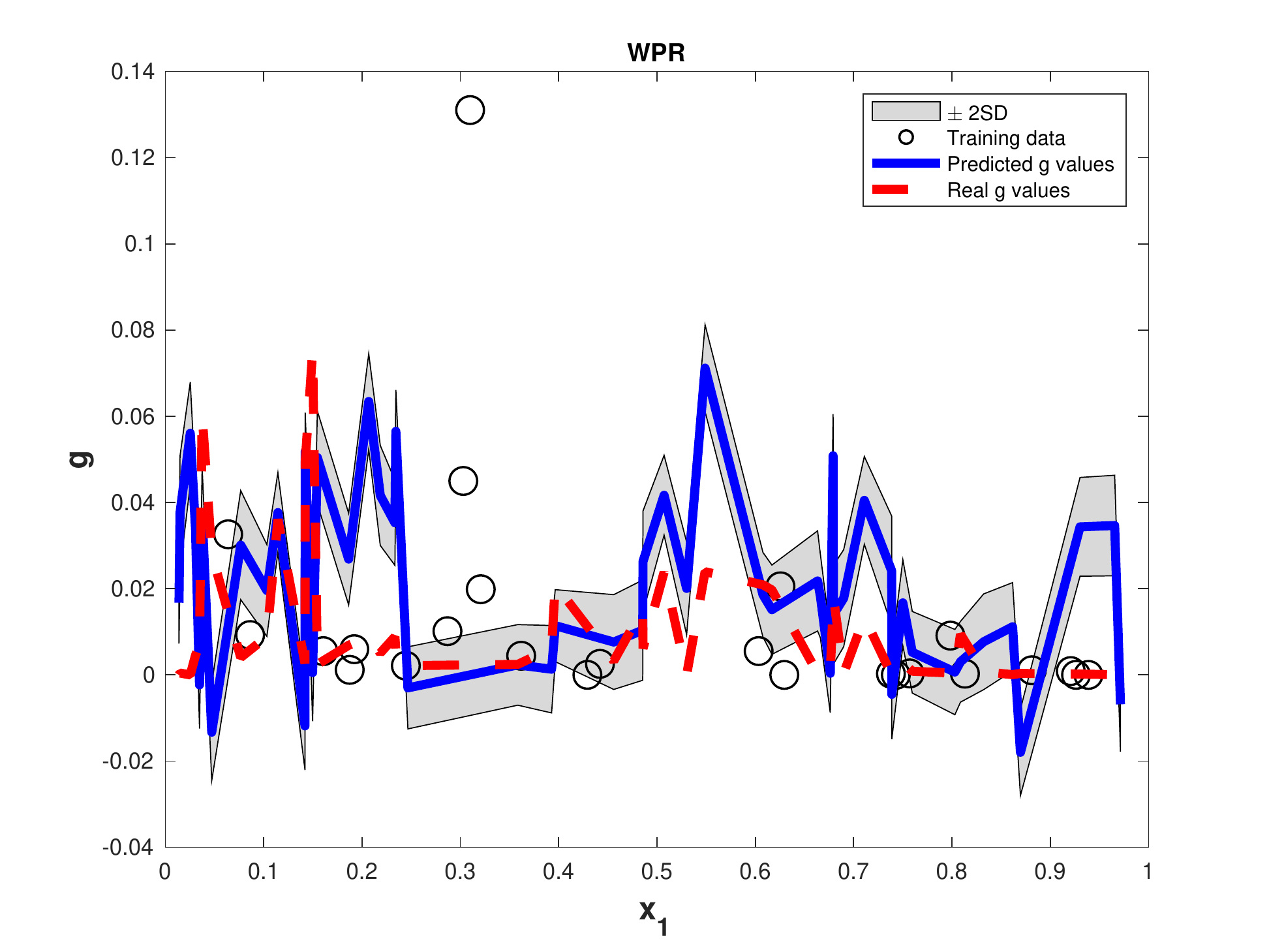}
    \includegraphics[scale=0.2]{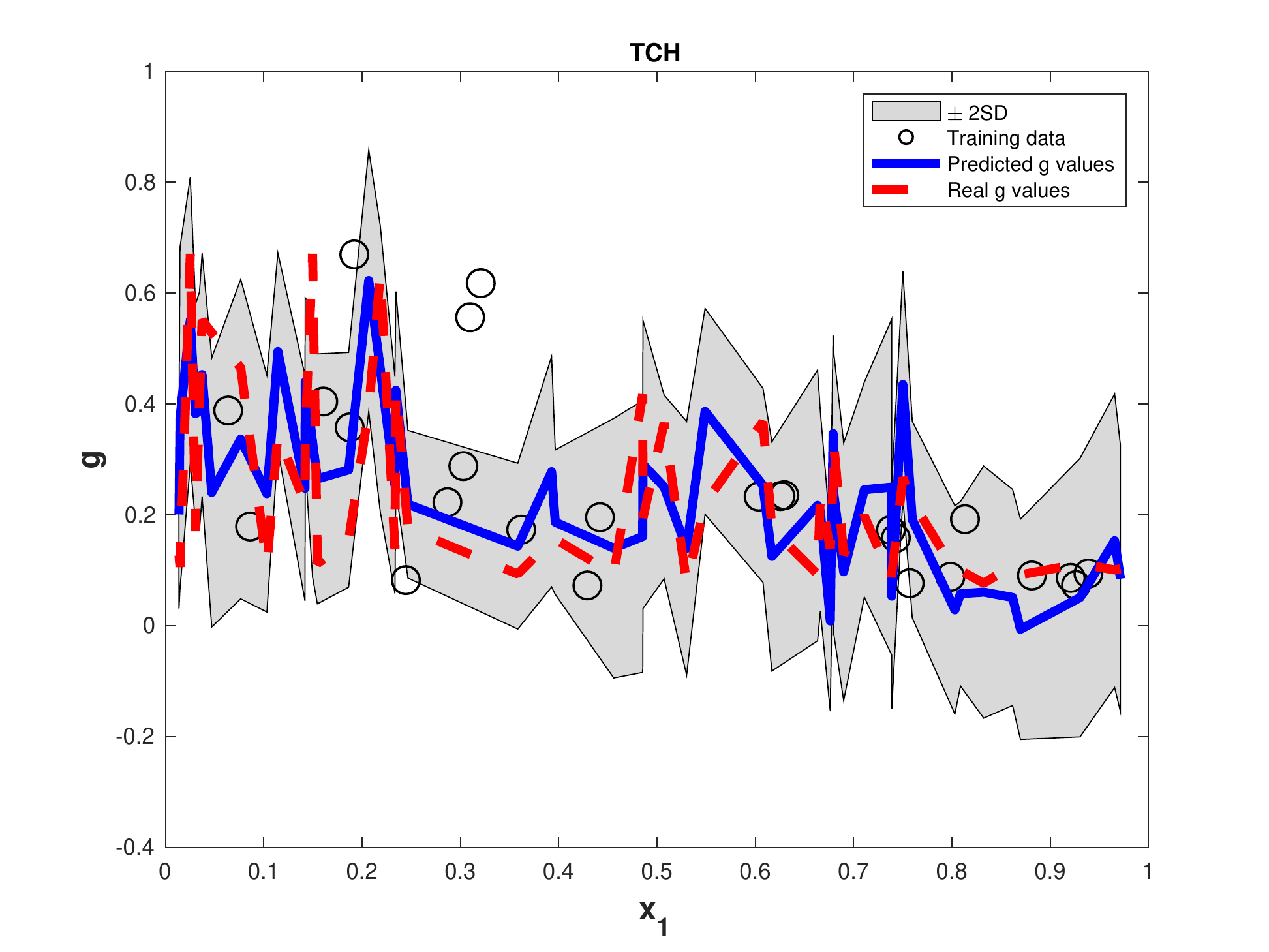}
    \includegraphics[scale=0.2]{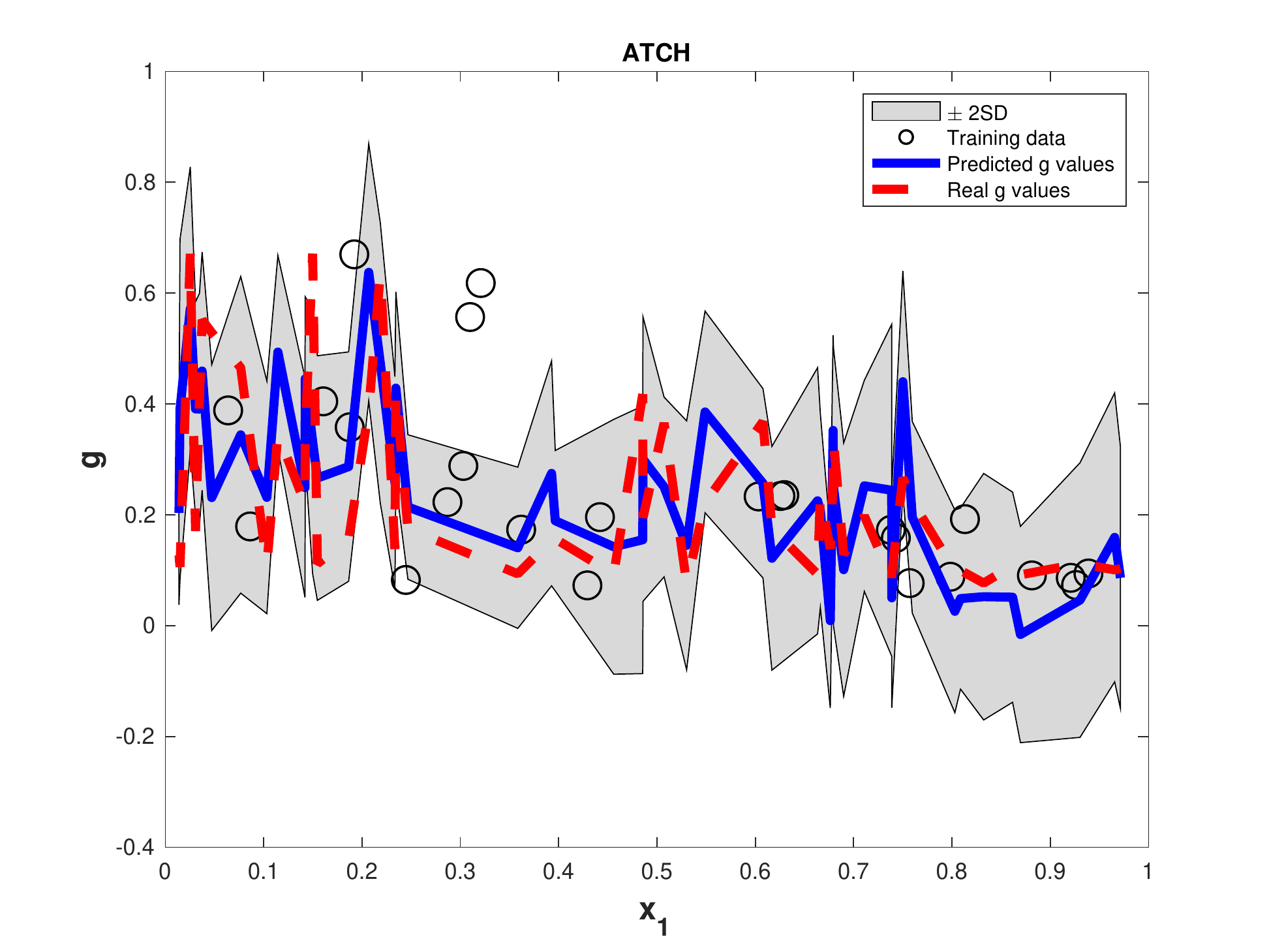}
    \includegraphics[scale=0.2]{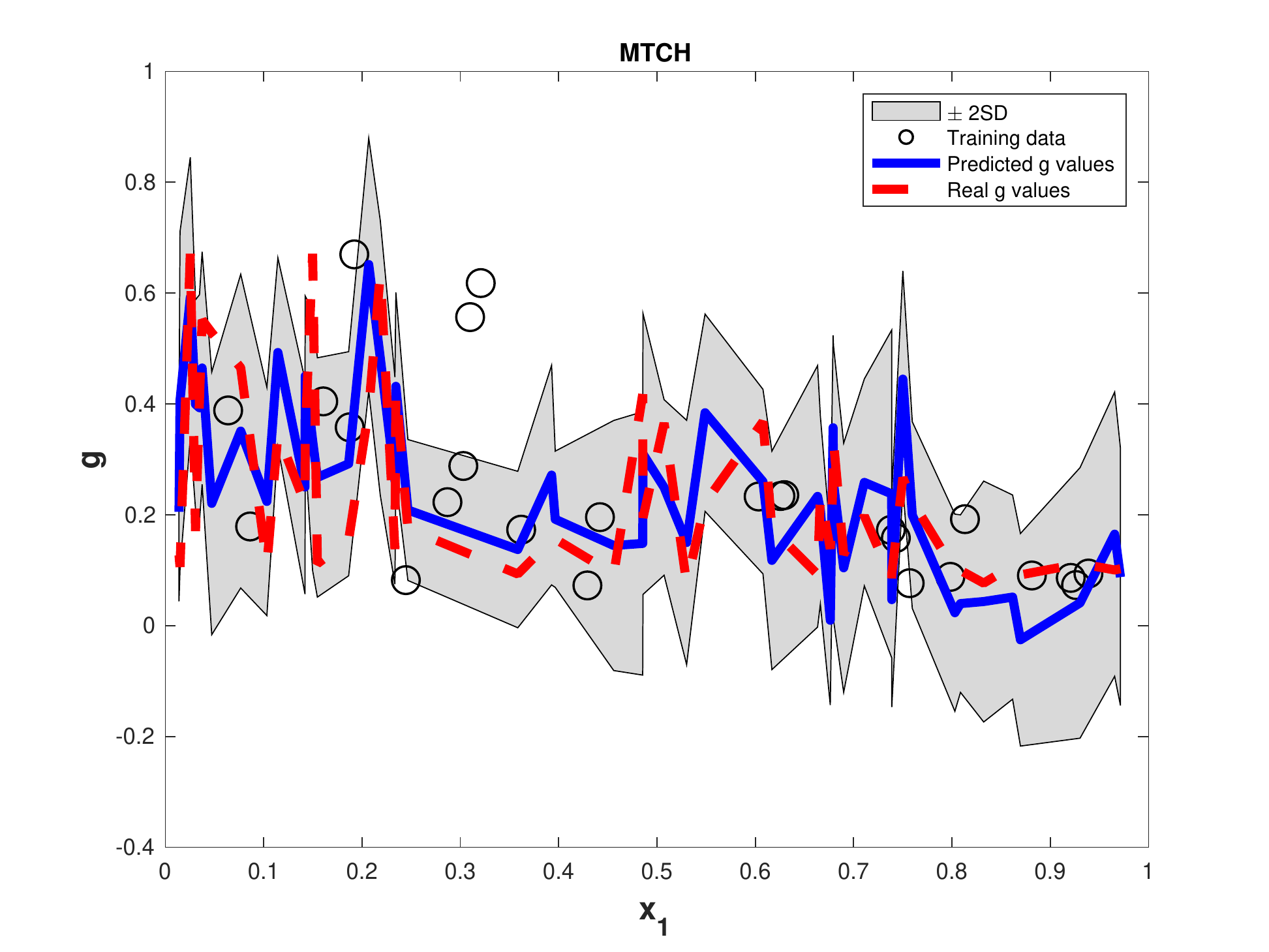}
    \includegraphics[scale=0.2]{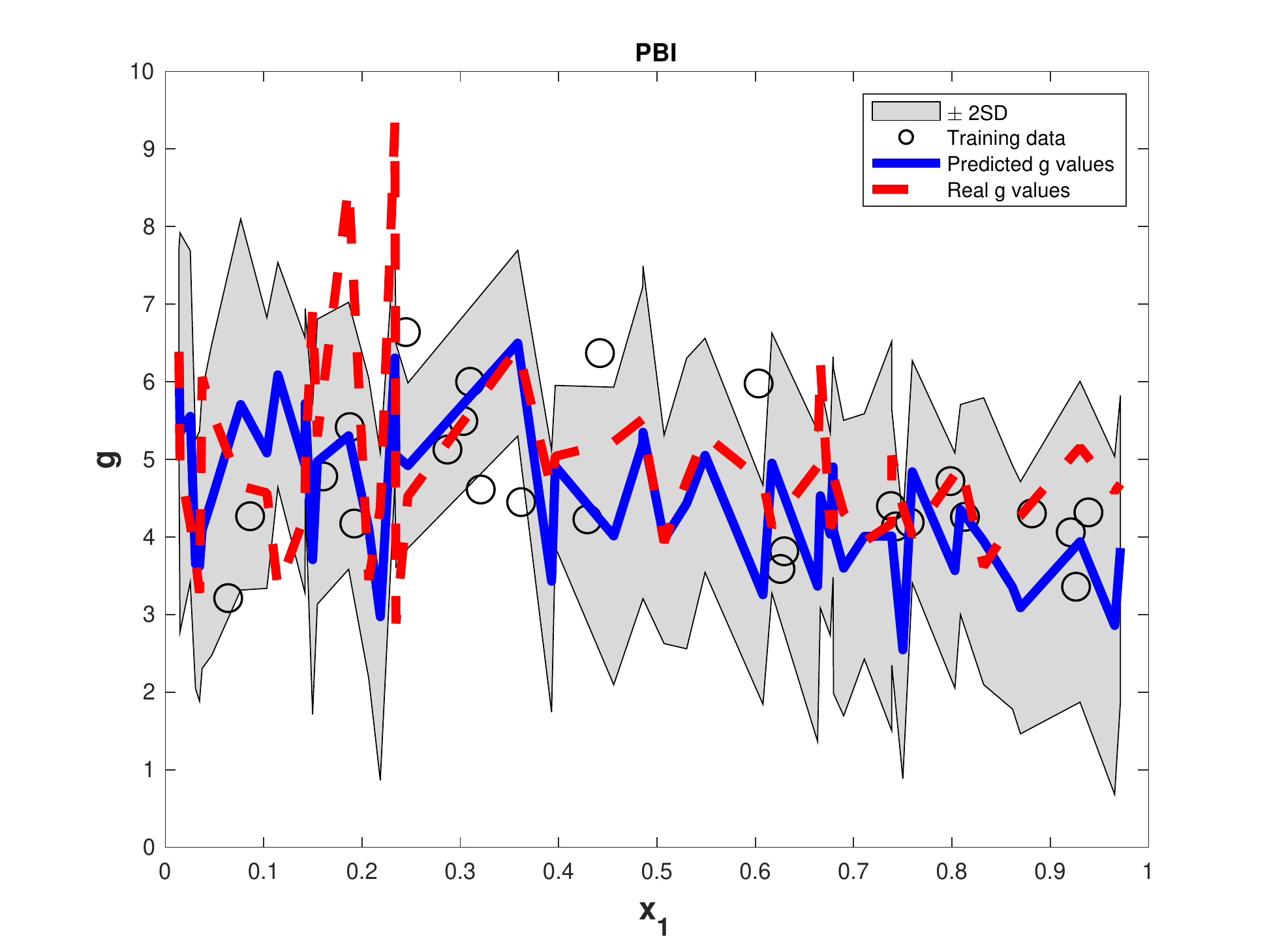}
    \includegraphics[scale=0.2]{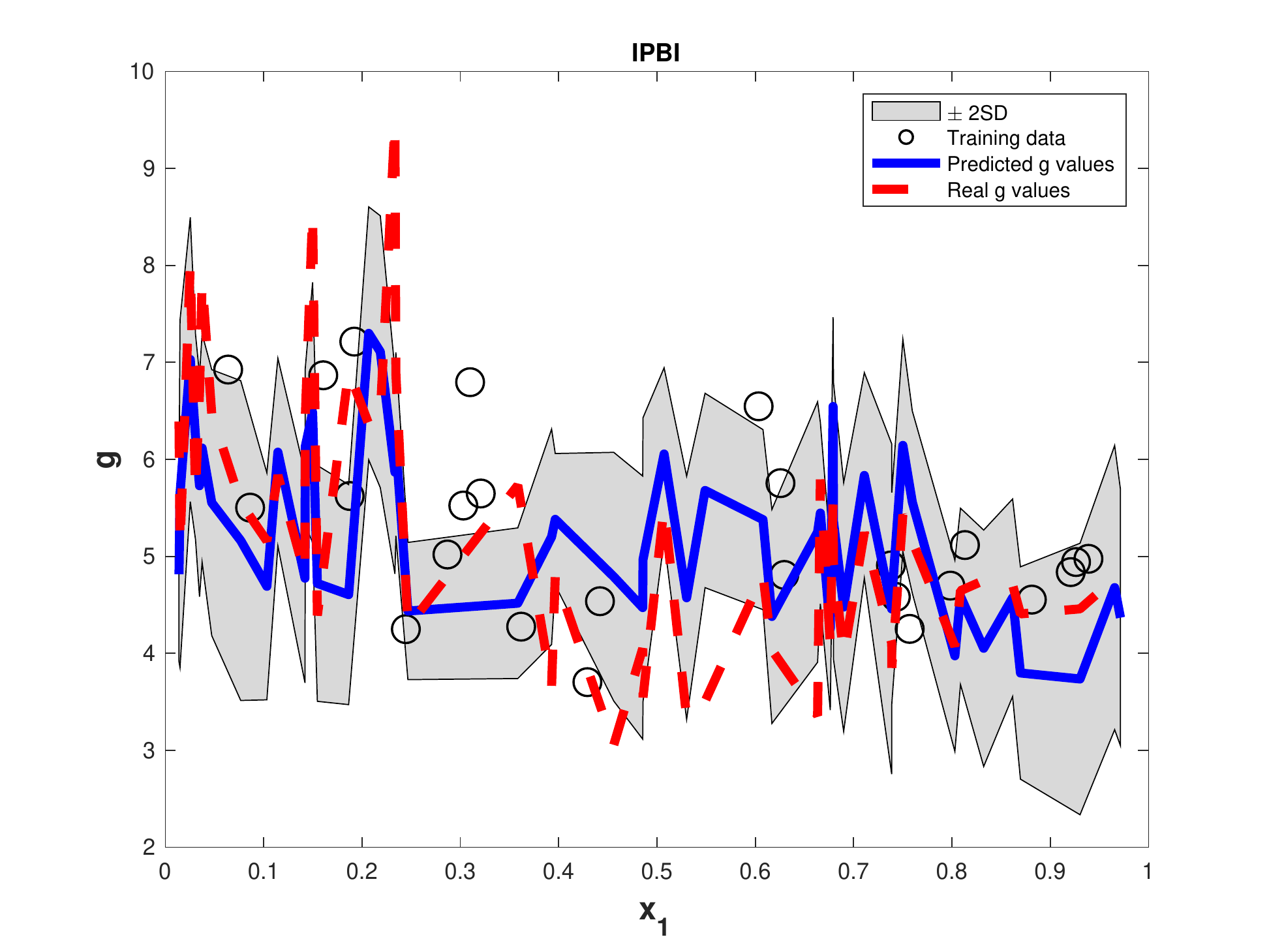}
    \includegraphics[scale=0.2]{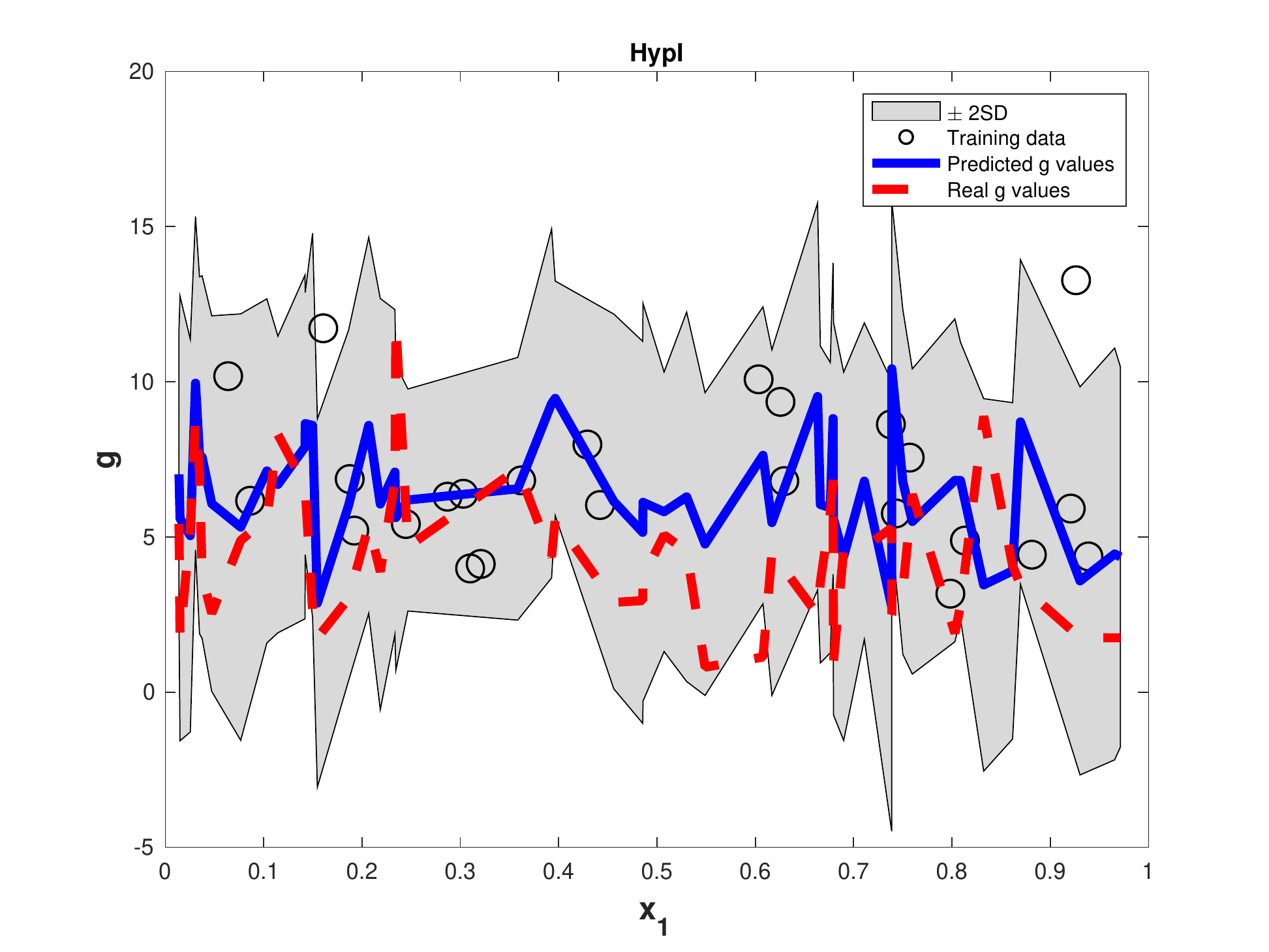}
    \includegraphics[scale=0.2]{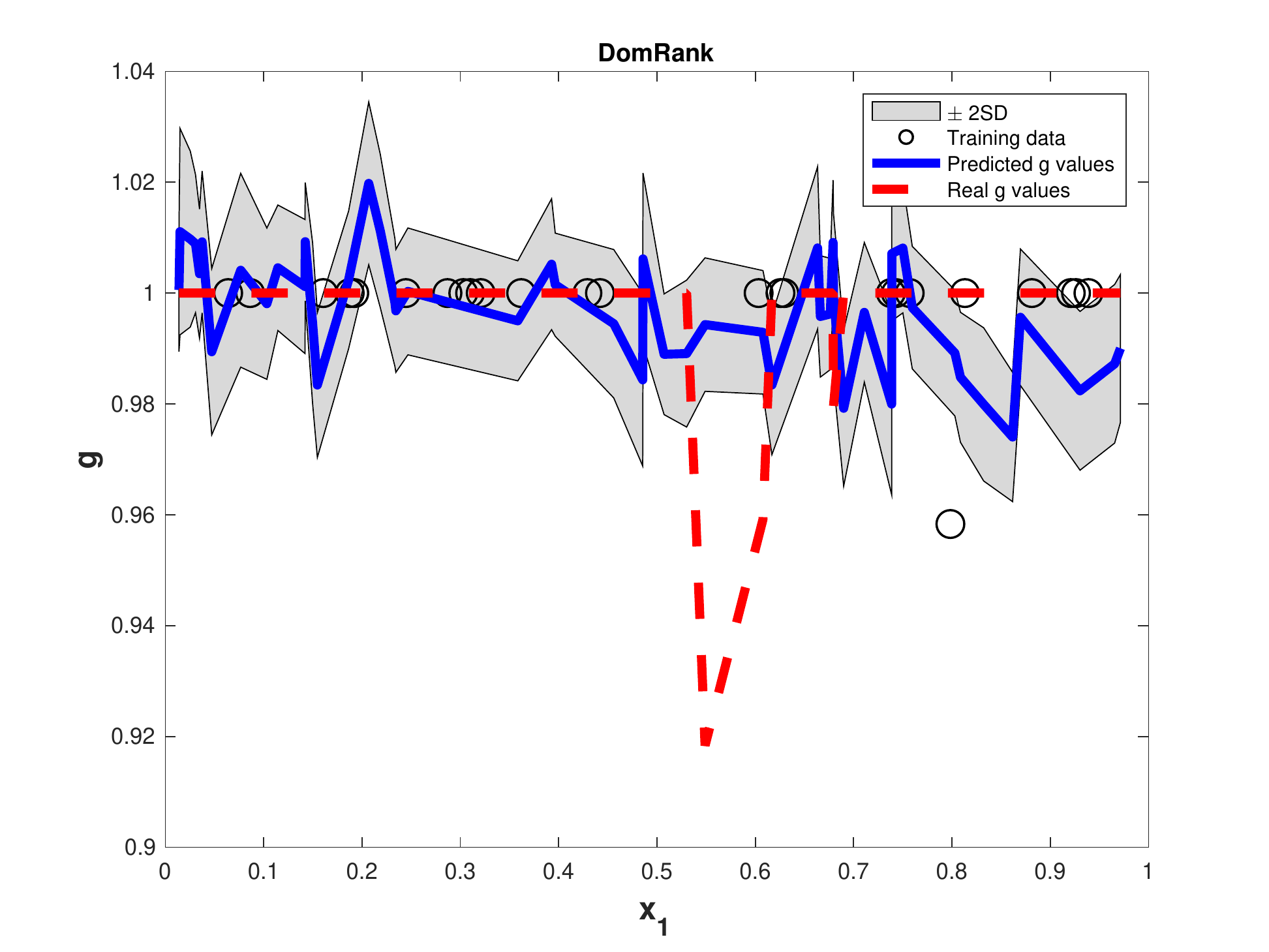}
    \includegraphics[scale=0.2]{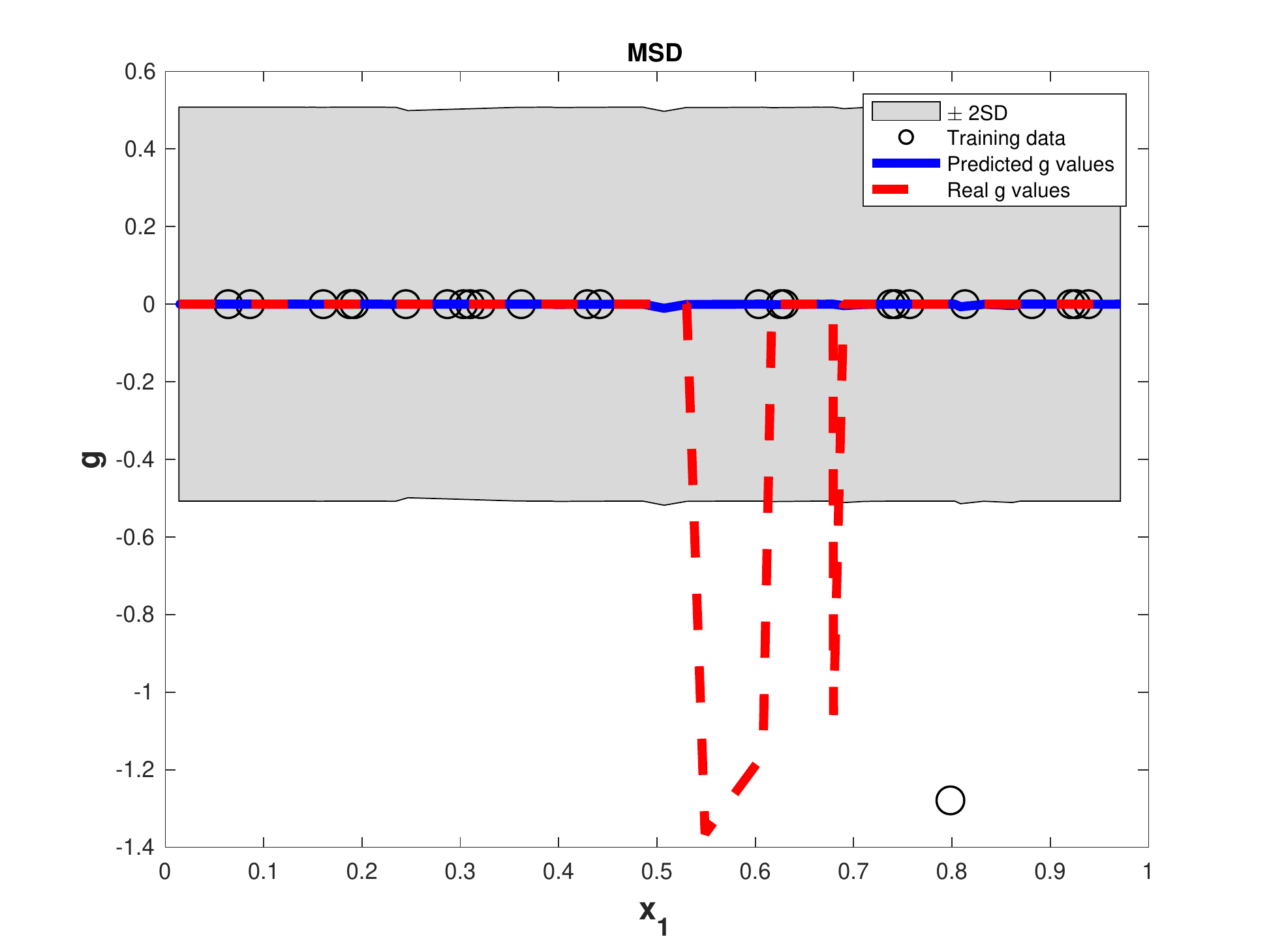}
    \includegraphics[scale=0.2]{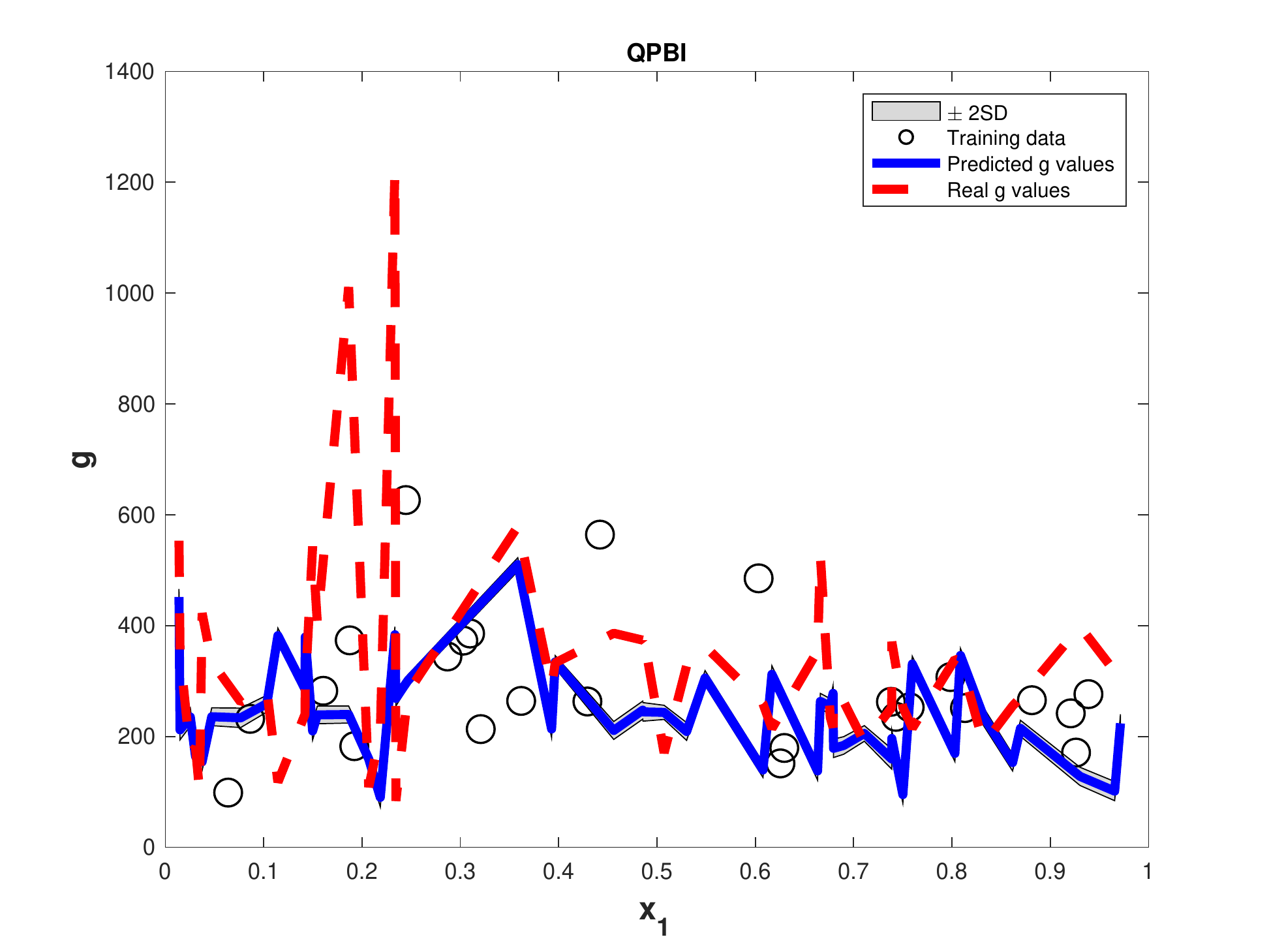}
    \includegraphics[scale=0.2]{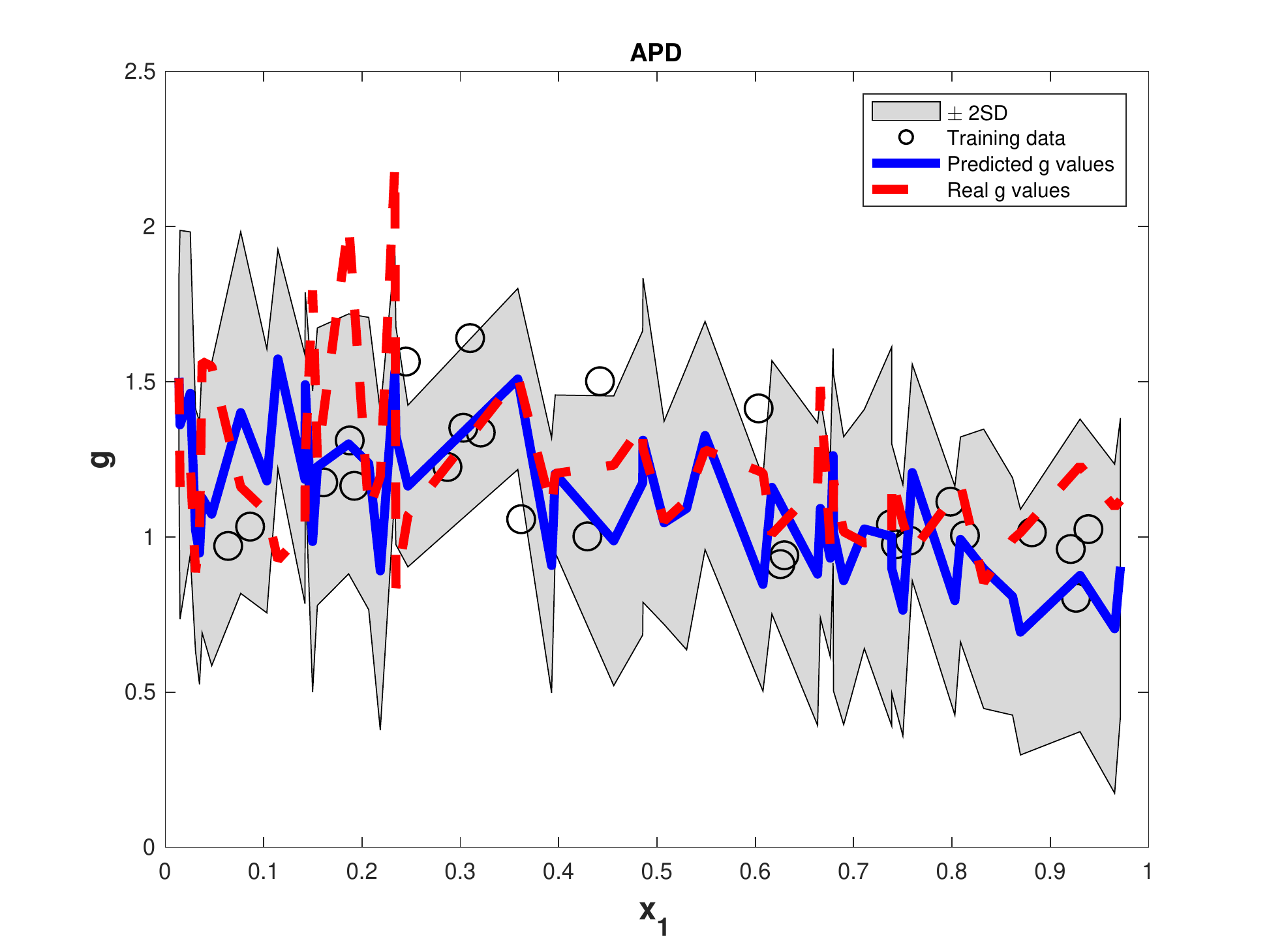}
    \caption{\label{fig:fitness_landscape_DTLZ2_10}Approximated scalarizing function values (notated by $g$) with one decision variable value for DTLZ2 10 objectives, $\pm $2SD represents the approximated $g$ values with $\pm$ 2 standard deviations}
\end{figure*}

As can be seen in the figure that the performance of different functions varies with the number of objectives. The reason is that different formulations of scalarizing functions provide a different landscape. For a better illustration, an example of fitness landscapes of DTLZ2 with two and 10 objectives for different functions are shown in Figures \ref{fig:fitness_landscape_DTLZ2_2} and \ref{fig:fitness_landscape_DTLZ2_10}, respectively. In figures, we used a training data set (shown as circles in figures) and built the surrogates on scalarizing functions. Then we generated a testing data set and used the model to get the predicted values (solid lines in figures) and uncertainties of the predicted values (shaded region in figures). The dotted line represents the actual scalarizing function values (denoted by g in figures) on the testing data set. For visualization purposes, we could show only one of the variables values.

As can be seen, for the same training data set (and therefore same objective function values), different scalarizing functions give different values. Therefore, the surrogate model to be built strongly depends on the g function values. For instance, in many-objective case, most of the solutions were nondominated and functions which use the property of Pareto dominance produce similar g values, as can be seen in the plots of HypI, DomRank, and MSD in 10 objective cases. As the g function values are the same for most of the solutions, the algorithm (EGO in this work) for finding a sample to update the surrogate by optimizing an infill criterion (EI in this work) will not be able to enhance the accuracy of the surrogates This is also the reason that HypI performed very well in a low number of objectives and its performance deteriorated with an increase in the number of objectives. On the other hand, functions like APD and PBI are not very sensitive to the number of objectives as they do not use the property of Pareto dominance and suitable for a large number of objectives when building surrogates on them. In addition to the sensitivity towards many objectives, some functions, EWC and WS did not perform well in most of the problems. In EWC, the g function landscape is very rugged \cite{Malan2009} as can be seen in figures and WS has the problem of finding solutions in the non-convex parts of the Pareto front. These results show that one needs to see or analyze the landscape of scalarizing functions for the given training data set before using them in the optimization. We observed the similar behavior on other problems and landscapes on other problems with different objectives are provided in the supplementary material.




\section{Conclusions and future research directions}
This work focused on reviewing, analyzing and comparing different scalarizing functions in Bayesian multiobjective optimization for solving (computationally) expensive multi- and many-objective optimization problems. We provided an overview of different functions with their merits and demerits. We built the surrogates on the different functions in the framework of Bayesian multiobjective optimization. We compared different functions with the different number of objectives by using IGD and hypervolume as the performance metrics. The results clearly showed that some functions outperformed others in many cases and some did not work in most of the cases. We then analyzed the fitness landscape of different functions with respect to the number of objectives. We found out that some of the functions are very sensitive to the number of objectives and an analysis of the landscape of different functions might be helpful in selecting functions when building surrogates on them and using them in the optimization for solving expensive MOPs.

In this work, we did not compare the scalarizing functions with other non-Bayesian multiobjective optimization algorithms. Therefore, comparison with other algorithms are topics for future research directions. Moreover, working on an adaptive strategy in using and selecting scalarizing functions and their corresponding parameters are also in the list of future research. 



\section*{Acknowledgement}
This research is supported by the Natural Environment Research Council [NE/P017436/1]


\bibliographystyle{plain}
\bibliography{Master_File}




\end{document}